%% file: main.tex
\documentclass[letterpaper]{article} 
\PassOptionsToPackage{table}{xcolor}
\usepackage[preprint]{aaai2027}  
\usepackage[hyphens]{url}  
\usepackage{graphicx} 
\usepackage{natbib}  
\usepackage{caption} 
\usepackage{algorithm}
\usepackage{algorithmic}
\usepackage{xspace}
\usepackage{amssymb}
\usepackage[most]{tcolorbox}
\usetikzlibrary{calc}

\usepackage{newfloat}
\usepackage{listings}
\DeclareCaptionStyle{ruled}{labelfont=normalfont,labelsep=colon,strut=off} 
\floatstyle{ruled}
\newfloat{listing}{tb}{lst}{}
\floatname{listing}{Listing}

\usepackage{booktabs}
\usepackage{multirow}
\usepackage{arydshln}
\input{Appendix/appendix_setup.tex}

\title{Self-Evolving Neuro-Symbolic Skills for Tool-Augmented Spatial Reasoning}
\author{
    Shi-Yu Tian\textsuperscript{\rm 1,\rm 2}\equalcontrib,
    Zhuo-Xia Wang\textsuperscript{\rm 1,\rm 2}\equalcontrib,
    Xuan-Yi Zhu\textsuperscript{\rm 1,\rm 2},
    Zhi Zhou\textsuperscript{\rm 1},\\
    Xinwei Yang\textsuperscript{\rm 3},
    Kun-Yang Yu\textsuperscript{\rm 1,\rm 2},
    Ming Yang\textsuperscript{\rm 1,\rm 2},
    Yang Chen\textsuperscript{\rm 1},
    Yu-Feng Li\textsuperscript{\rm 1,\rm 2}\corresponding
}
\affiliations{
    \textsuperscript{\rm 1}National Key Laboratory for Novel Software Technology, Nanjing University, China\\
    \textsuperscript{\rm 2}School of Artificial Intelligence, Nanjing University, China\\
    \textsuperscript{\rm 3}College of Computer Science, Sichuan University
}

\newcommand{\method}{\textsc{NeSy-Spatial}\xspace}

\newtcolorbox[auto counter, number freestyle={\noexpand\arabic{\tcbcounter}}]{definedbox}[2][]{%
    enhanced,
    colback=black!5!white,
    colframe=black!75!white,
    title=#2,
    #1
}

\newtcolorbox{formalizedprompt}[1][]{%
  enhanced,
  breakable,
  colback=gray!5,
  colframe=gray!80,
  fonttitle=\bfseries,
  title=Formalize2Tabular\_prompt,
  coltitle=black,
  sharp corners,
  boxrule=0.8pt,
  arc=2mm,
  left=4pt,
  right=4pt,
  top=6pt,
  bottom=6pt,
  width=\textwidth,
  #1
}

\begin{document}

\maketitle

\begin{abstract}

Large vision-language models have achieved strong performance in multimodal reasoning, but they remain unreliable on fine-grained spatial tasks that demand both precise spatial perception and fine-grained geometric computation beyond end-to-end generation. Tool augmentation offers a natural solution, while existing methods either plan tool calls from scratch without explicit dependency constraints or rely on fixed pipelines that are redundant and generalize poorly across spatial tasks. An effective spatial reasoning agent should instead accumulate reusable experience and adaptively compose it for new problems.
To this end, we propose \method, a neuro-symbolic framework for self-evolving spatial skills. \method abstracts tool interactions and geometric operations into typed executable atomic instructions and composes them into two complementary skill types: \textit{Tool-Use Skills} for organizing tool execution and \textit{Geometry Skills} for structured geometric reasoning. During inference, \method retrieves and executes relevant skills in a closed-loop process. During evolution, it analyzes buffered successful and failed trajectories to refine skill structures and prune unreliable or inactive entries. Experiments on three spatial reasoning benchmarks show that \method consistently improves reasoning accuracy with more precise tool utilization.

\end{abstract}


\input{01_intro.tex}
\input{02_relatedwork.tex}
\input{03_methods.tex}
\input{04_experiments.tex}
\input{05_conclusion.tex}

\bibliography{aaai2027}

\input{Appendix/appendix.tex}


\end{document}

%% file: Appendix/appendix_setup.tex
\usepackage{amsmath}
\usepackage{pifont}
\usepackage{tabularx}
\usepackage{array}
\usepackage{xurl}
\usepackage{enumitem}

\newcolumntype{Y}{>{\centering\arraybackslash}X}

\definecolor{cardback}{RGB}{252,252,252}
\definecolor{cardframe}{RGB}{155,155,155}
\definecolor{cardtitle}{RGB}{211,211,211}
\definecolor{cardsubtle}{RGB}{241,241,241}
\definecolor{promptback}{RGB}{249,249,249}
\definecolor{prompttitle}{RGB}{198,198,198}

\lstdefinestyle{jsonplain}{
    basicstyle=\ttfamily\fontsize{7.05}{7.75}\selectfont,
    breaklines=true,
    breakatwhitespace=false,
    columns=fullflexible,
    keepspaces=true,
    showstringspaces=false,
    frame=none,
    backgroundcolor=\color{white},
    xleftmargin=2pt,
    xrightmargin=2pt,
    aboveskip=1pt,
    belowskip=1pt,
    lineskip=-0.3pt
}

\lstdefinestyle{pythonplain}{
    language=Python,
    basicstyle=\ttfamily\fontsize{7.05}{7.75}\selectfont,
    breaklines=true,
    breakatwhitespace=false,
    columns=fullflexible,
    keepspaces=true,
    showstringspaces=false,
    frame=none,
    backgroundcolor=\color{white},
    xleftmargin=2pt,
    xrightmargin=2pt,
    aboveskip=1pt,
    belowskip=1pt,
    lineskip=-0.3pt
}

\newcommand{\casestep}[1]{%
  \par\medskip
  \noindent{\small\bfseries #1}\enspace
  \ignorespaces
}

\tcbuselibrary{breakable,listings}

\newtcolorbox{promptbox}[1]{
  breakable,
  colback=promptback,
  colframe=cardframe,
  colbacktitle=prompttitle,
  coltitle=black,
  fonttitle=\bfseries\small,
  title={#1},
  boxrule=0.75pt,
  arc=3pt,
  outer arc=3pt,
  left=6pt,right=6pt,top=5pt,bottom=5pt,
  before skip=7pt,
  after skip=7pt
}

\newtcolorbox{skillcard}[1]{
  breakable,
  colback=cardback,
  colframe=cardframe,
  colbacktitle=cardtitle,
  coltitle=black,
  fonttitle=\bfseries\small,
  title={#1},
  boxrule=0.6pt,
  arc=2pt,
  outer arc=2pt,
  left=6pt,right=6pt,top=5pt,bottom=5pt,
  before skip=7pt,
  after skip=7pt,
  fontupper=\small
}

\newcommand{\skillid}[1]{%
  \noindent\textbf{Internal ID.}\enspace
  {\ttfamily\footnotesize\path{#1}}\par\smallskip
}

\newtcolorbox{questionbox}{
  breakable,
  colback=cardsubtle,
  colframe=cardframe,
  boxrule=0pt,
  leftrule=2pt,
  arc=0pt,
  left=6pt,right=5pt,top=4pt,bottom=4pt,
  before skip=5pt,
  after skip=6pt,
  fontupper=\small
}

\newtcolorbox{resultbox}{
  breakable,
  colback=cardsubtle,
  colframe=cardframe,
  boxrule=0.5pt,
  arc=1.5pt,
  left=6pt,right=6pt,top=4pt,bottom=4pt,
  before skip=5pt,
  after skip=5pt,
  fontupper=\small
}

\newtcblisting[auto counter]{jsoncard}[2]{
  breakable,
  listing only,
  listing engine=listings,
  colback=white,
  colframe=cardframe,
  colbacktitle=cardtitle,
  coltitle=black,
  fonttitle=\bfseries\small,
  title={Listing~\thetcbcounter: #1},
  label={#2},
  boxrule=0.6pt,
  arc=2pt,
  outer arc=2pt,
  left=3pt,right=3pt,top=2pt,bottom=2pt,
  before skip=7pt,
  after skip=7pt,
  listing options={style=jsonplain}
}

\newtcblisting[use counter from=jsoncard]{pythoncard}[2]{
  breakable,
  listing only,
  listing engine=listings,
  colback=white,
  colframe=cardframe,
  colbacktitle=cardtitle,
  coltitle=black,
  fonttitle=\bfseries\small,
  title={Listing~\thetcbcounter: #1},
  label={#2},
  boxrule=0.6pt,
  arc=2pt,
  outer arc=2pt,
  left=3pt,right=3pt,top=2pt,bottom=2pt,
  before skip=7pt,
  after skip=7pt,
  listing options={style=pythonplain}
}

\newtcblisting{codecard}[1]{
  breakable,
  listing only,
  listing engine=listings,
  colback=white,
  colframe=cardframe,
  colbacktitle=cardtitle,
  coltitle=black,
  fonttitle=\bfseries\small,
  title={#1},
  boxrule=0.6pt,
  arc=2pt,
  outer arc=2pt,
  left=3pt,right=3pt,top=2pt,bottom=2pt,
  before skip=7pt,
  after skip=7pt,
  listing options={style=pythonplain}
}

%% file: 01_intro.tex
\section{Introduction}

Large vision-language models (LVLMs) have made substantial progress in general visual understanding~\cite{GPT4}, yet they remain unreliable for fine-grained spatial reasoning~\cite{yang2025vsibench,chen2025thinkwith3d_limitedview}. Unlike generic visual understanding, spatial reasoning often requires explicit geometric reasoning and intermediate verification beyond end-to-end generation. Such reasoning is difficult to accomplish using LVLMs alone, as it depends on precise visual perception and structured geometric computation. Tool augmentation has therefore emerged as an effective solution~\cite{zhang2025thyme,yu2026thinkingwithtable,tian2026last}, enabling LVLMs to leverage specialized vision models and external executors such as depth estimators~\cite{depthanythingv2} or reconstruction models~\cite{Wang_2025_vggt} to obtain reliable intermediate evidence for reasoning.

\begin{figure}[t]
    \centering
    \includegraphics[width=\columnwidth]{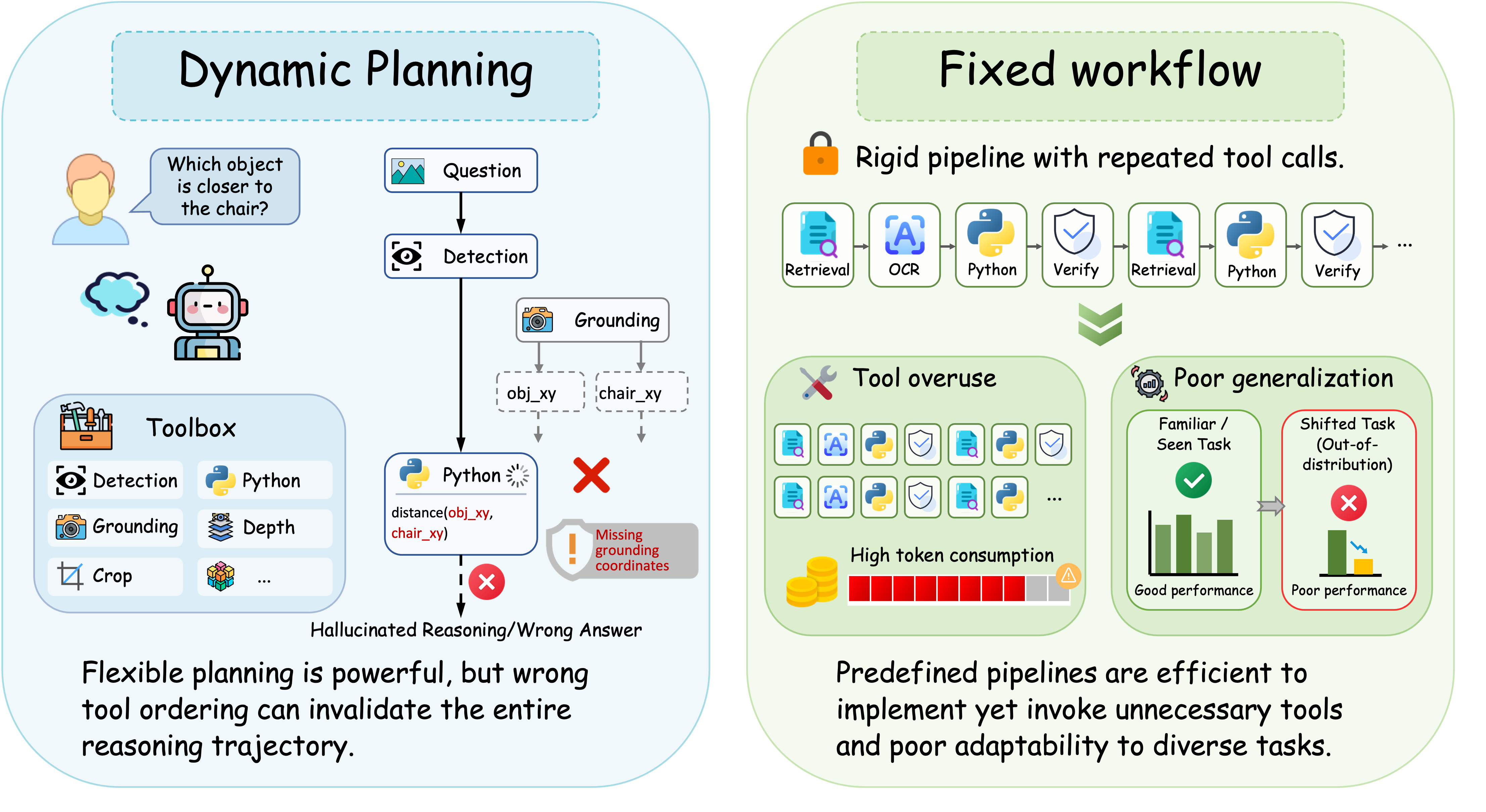}
    \caption{Limitations of tool-augmented spatial reasoning agents: dynamic planning may violate tool dependencies, while fixed pipelines can be redundant and generalize poorly.}
    \label{fig:intro_motivation}
\end{figure}

However, existing tool-augmented methods mainly follow two paradigms. One line of work dynamically plans tool calls from a general-purpose toolbox for each query~\cite{tian2026last,wu2026spatialscore}, while another executes a manually designed pipeline tailored to a specific class of spatial reasoning tasks~\cite{cho2026spatialclaw,chen2026gca}. Although dynamic planning is highly flexible, it lacks explicit constraints on tool dependencies and execution order. As a result, agents may generate invalid tool sequences—for example, invoking a Python interpreter before acquiring prerequisite information, such as grounded object coordinates or point cloud representations, causing subsequent reasoning to fail. In contrast, fixed pipelines enforce a valid execution order but often invoke unnecessary tools regardless of task complexity. Even simple queries may trigger the entire workflow, leading to redundant computation and reduced efficiency, while handcrafted pipelines are difficult to generalize across diverse spatial reasoning scenarios. 

Therefore, a spatial reasoning agent should abstract retrievable and composable skills from historical trajectories, rather than planning from scratch or relying on fixed templates. However, this goal presents two challenges. First, spatial skills involve both tool invocation and geometric processing. Existing methods typically rely on a Python interpreter to perform ad hoc computations over tool outputs, resulting in limited robustness and success rates. Second, skills are difficult to extract directly from trajectories, as successful trajectories may contain redundant steps, while failed ones may still include locally effective operations. The key challenge is therefore to identify reusable atomic instructions and compose them into high-level skills that jointly support tool execution and geometric reasoning.

To overcome this, we propose \method, a neuro-symbolic framework for self-evolving skills in spatial reasoning. Rather than simply expanding the available toolset, \method extracts atomic instructions from spatial reasoning trajectories and further composes them into reusable high-level skills. Specifically, we define two complementary types of skills: \textit{Tool-Use Skill} and \textit{Geometry Skill}. \textit{Tool-Use Skills} organize the execution of external tools, where atomic instructions correspond to individual tool calls and high-level skills encode their execution order and dependency structure. \textit{Geometry Skills} structure geometric reasoning, where atomic instructions correspond to executable functional code blocks and high-level skills abstract reusable operations such as coordinate transformation, distance computation, and consistency checking.

\method continuously learns and abstracts skills during online interaction, and further leverages the evolving skill library to guide subsequent reasoning. This process consists of two stages: inference and evolution. During inference, \method retrieves, instantiates, and executes relevant skills in a closed-loop process, dynamically adapting subsequent reasoning based on intermediate results. During evolution, the model mines new atomic instructions and local rules from both successful and failed execution trajectories, integrates them with existing skills, and removes unreliable or inactive entries according to verification results and usage patterns. Experiments on three datasets show that \method consistently improves spatial reasoning accuracy while achieving more precise tool utilization.

Our contributions are summarized as follows:
\begin{itemize}
    \item We introduce a neuro-symbolic definition of spatial skills that unifies tool invocation and geometric computation through typed executable atomic instructions.
    \item We propose \method, a skill self-evolution framework that updates atomic instructions and high-level Tool-Use and Geometry Skills from buffered trajectories through analysis, fusion, and pruning.
    \item We evaluate \method on multiple spatial reasoning benchmarks, showing improvements over base static tool-use agents and fixed-pipeline methods in performance.
\end{itemize}

%% file: 02_relatedwork.tex
\begin{figure*}[t]
    \centering
    \includegraphics[width=\textwidth]{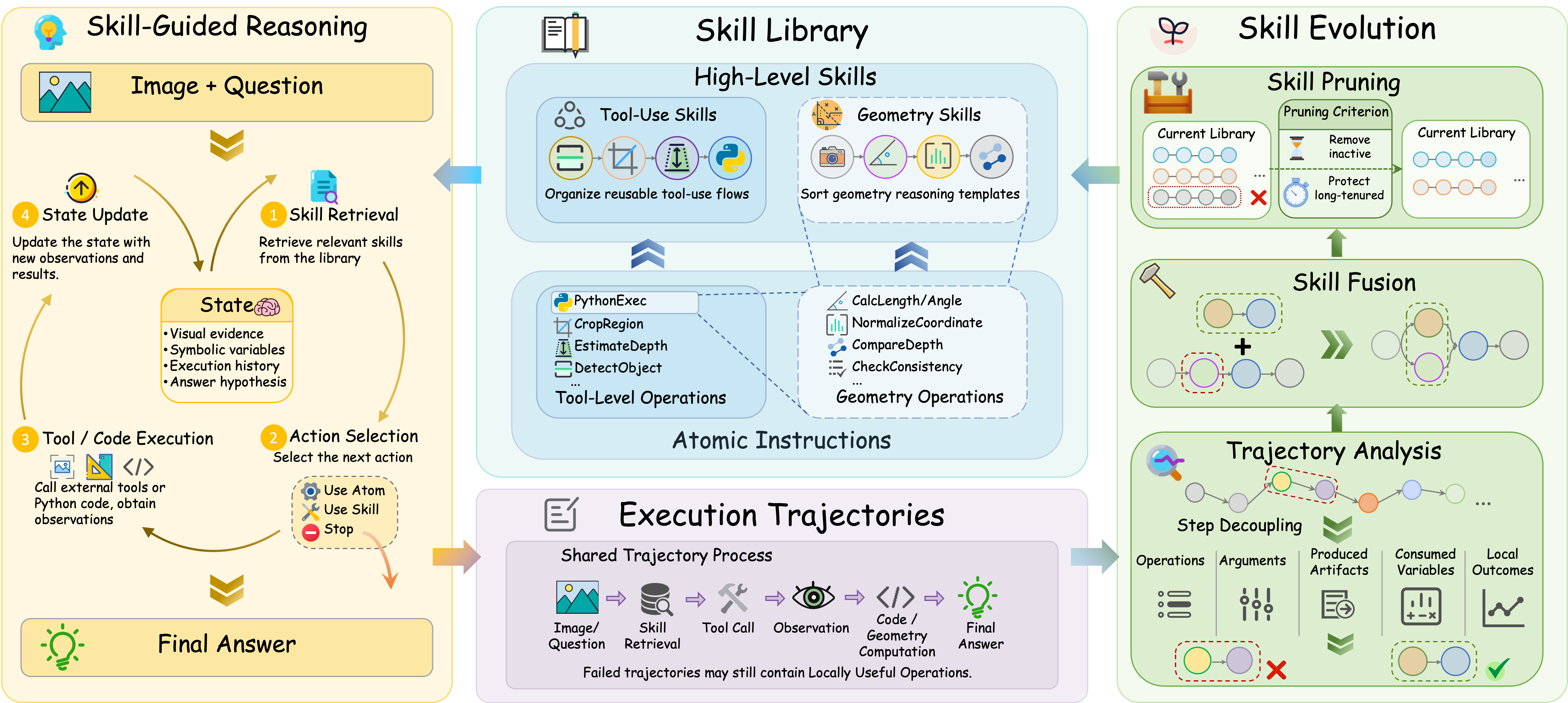}
    \caption{Overview of \method. \textbf{Left:} the agent retrieves and executes skills for closed-loop spatial reasoning. \textbf{Center top:} the Skill Library organizes atomic instructions into reusable high-level skills. \textbf{Center bottom:} execution trajectories record the shared reasoning process. \textbf{Right:} trajectory analysis, skill fusion, and pruning enable continual self-evolution.}

    \label{fig:framework}
\end{figure*}

\begin{figure*}[t]
    \centering
    \resizebox{\textwidth}{!}{\includegraphics{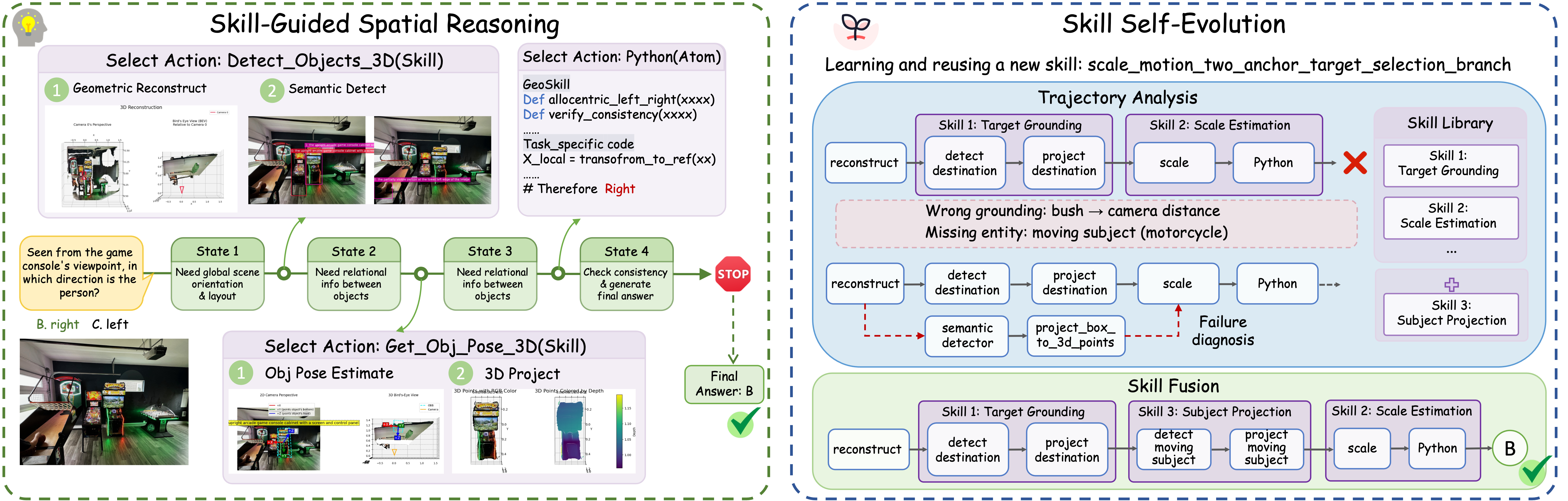}}
    \caption{Qualitative example of skill-guided spatial reasoning and skill evolution. \textbf{Left:} \method composes visual and geometric tools to solve a perspective-taking question. \textbf{Right:} failure diagnosis identifies missing grounding dependencies, updates the Tool-Use Skill, and enables the revised pipeline to produce the correct answer on a subsequent sample.}
    \label{fig:example}
\end{figure*}

\section{Related Work}

\subsection{Spatial Reasoning}
Spatial reasoning is the ability to interpret geometric relationships between objects and their environment~\cite{chen2024spatialvlm, liu2023visualspatial, yang2025vsibench}. It is fundamental to robotic perception and embodied intelligence~\cite{liu2025embodiedsurvey}, as well as autonomous driving~\cite{chen2024autodriving}. Recent benchmarks extend earlier static settings to multi-view observations, viewpoint changes, scale estimation, and dynamic scenes, including BLINK~\cite{fu2024blink}, CVBench~\cite{tong2024cvbench}, VSI-Bench~\cite{yang2025vsibench}, and STI-Bench~\cite{li2025stibench}. Our evaluation covers MMSI~\cite{yang2025mmsi}, MindCube~\cite{chen2026mindcube}, and OmniSpatial~\cite{jia2025omnispatial}. Despite advances in general perception, vision-language models still struggle with multi-step spatial inference, precise scale estimation~\cite{chen2025sdvlm_msmu}, and viewpoint-dependent reasoning~\cite{li2025viewspatial}.

Existing solutions include task-specific fine-tuning~\cite{cheng2024spatialrgpt}, progressive training, 3D-aware representations~\cite{xu2024pointllm}, and modular systems that invoke visual tools for detection, depth estimation, reconstruction, or geometric measurement~\cite{chen2025spacetools, han2025tigertoolintegratedgeometricreasoning}. Recent tool-augmented agents include SpatialScore's SpaAge-PE and SpaAge-ReAct~\cite{wu2026spatialscore}, SpatialCLAW~\cite{cho2026spatialclaw}, GCA~\cite{chen2026gca}, and LAST~\cite{tian2026last}. Although explicit visual and geometric operations improve spatial reasoning, these methods primarily optimize individual reasoning pipelines rather than accumulating, reusing, and refining procedures across tasks.

\subsection{Skill Self-evolution}
Skill self-evolution enables agents to abstract reusable knowledge from past interactions to improve future task solving. Reflection-based methods revise reasoning through feedback, self-critique, or external validation~\cite{shinn2023reflexion, madaan2023selfRefine, gou2024critic}, while experience-driven methods retrieve knowledge abstracted from historical trajectories~\cite{zhao2024expel,shao2026nsi}. Skill-memory systems further maintain reusable skills or workflows to guide future actions~\cite{wang2023voyager, wang2024agent, wang2025inducing}. Together, these studies show that agent experience can be accumulated and reused.
However, prior studies largely address text reasoning, web navigation, coding, or games, representing skills as instructions, workflows, or programs. Tool-augmented spatial reasoning instead requires coordination among visual perception, geometric computation, and multi-step inference. It therefore calls for reusable procedures that specify when to invoke tools, how to interpret their outputs, and how to transfer successful geometric reasoning across spatial tasks.

%% file: 03_methods.tex
\section{Method}


\subsection{Problem Setup}
We study online tool-augmented spatial reasoning. At episode $t$, the agent receives $x_t=(I_t,q_t)\in\mathcal{I}\times\mathcal{Q}$ and predicts $\hat{y}_t\in\mathcal{Y}$ using a tool set $\mathcal{T}=\mathcal{T}_{\mathrm{vis}}\cup\mathcal{T}_{\mathrm{py}}$ and a skill library $\mathrm{Lib}_t=(\mathcal{A}_t,\mathcal{S}_t)$. The Python executor is exposed as a single tool-level atomic instruction, while its internal geometric computation is constructed from reusable Geometry Skills. At reasoning step $k$, the working memory $M_{t,k}$ maintains the accumulated evidence, symbolic variables, execution history, verifier outcomes, and answer hypothesis.

During an episode, $\mathrm{Lib}_t$ is fixed. Under our test-then-update protocol, each prediction is recorded exactly once before its ground-truth answer is revealed; the label is then used only to evaluate that prediction and assess the completed trajectory $\tau_t$, which is appended to the trajectory buffer $\mathcal{B}_t$. Once the buffer reaches $N_{\mathrm{buf}}$, self-evolution extracts and integrates reusable structures and prunes unreliable or inactive entries to obtain $\mathrm{Lib}_{t+1}$; otherwise, the library remains unchanged. Thus, all library updates occur between episodes, as illustrated in Figure~\ref{fig:framework}.

\subsection{Neuro-Symbolic Skill Formulation}

Spatial reasoning must couple learned perception with verifiable geometric computation. At a fixed episode, we omit the time index and write the skill library as
\begin{equation}
\mathrm{Lib}=\bigl(\mathcal{A},\mathcal{S}\bigr),
\qquad
\mathcal{S}=\mathcal{S}_{\mathrm{tool}}\cup\mathcal{S}_{\mathrm{geo}},
\end{equation}
where $\mathcal{A}$ is a set of executable atomic instructions, while $\mathcal{S}_{\mathrm{tool}}$ and $\mathcal{S}_{\mathrm{geo}}$ contain Tool-Use and Geometry Skills, respectively. Both skill types are represented by atomic-action nodes, ordering edges, and an instantiation context.

\subsubsection{Atomic Instructions}

An atomic instruction is a typed executable operation with a local verifier:
\begin{equation}
a=\bigl(e_a,v_a\bigr),
\qquad
e_a:\mathcal{U}_{a}\rightarrow\mathcal{O}_{a},
\qquad
v_a:\mathcal{O}_{a}\rightarrow\{0,1\}.
\end{equation}
Here $e_a$ maps atom-specific typed inputs $\mathcal{U}_{a}$ to outputs $\mathcal{O}_{a}$, and $v_a$ validates the output. We partition $\mathcal{A}$ into controller-visible tool atoms $\mathcal{A}_{\mathrm{tool}}$ and internal geometry atoms $\mathcal{A}_{\mathrm{geo}}$. The former wraps visual tools and the Python executor, each as one atomic instruction; the latter contains reusable code blocks for geometric transformation, numerical comparison, and relation verification. Geometry atoms are executed inside the Python tool rather than selected as independent controller actions.

\subsubsection{High-Level Skills}
We represent each high-level skill as a node--edge--context structure:
\begin{equation}
s=\bigl(V,E,\kappa\bigr),
\qquad
V\subseteq\mathcal{A},
\quad
E\subseteq V\times V,
\end{equation}
where $V$ contains atomic instructions, $E$ specifies their execution dependencies, and $\kappa$ stores the context required for instantiation. Tool-Use and Geometry Skills share this representation but differ in their node types and execution roles.

\input{Figures/skill_examples}


A \textit{Tool-Use Skill} $s\in\mathcal{S}_{\mathrm{tool}}$ organizes controller-visible tool atoms into a reusable workflow. Its nodes satisfy $V\subseteq\mathcal{A}_{\mathrm{tool}}$, its edges define execution dependencies, and its context stores object bindings, tool arguments, and available evidence.
Python is treated as an atomic tool node to keep the controller action space compact, while its internal computation is handled separately by  \textit{Geometry Skills}.


A \textit{Geometry Skill} $s\in\mathcal{S}_{\mathrm{geo}}$ provides a reusable and verifiable implementation for a Python tool call. Its nodes satisfy $V\subseteq\mathcal{A}_{\mathrm{geo}}$, its edges encode data dependencies, and its context records reference frames, units, predicates, and numerical evidence.
When invoking a Python tool, the model retrieves relevant Geometry Skills to guide code construction before submitting the generated code to the interpreter.
Figure~\ref{fig:skill_examples} illustrates representative skills and their organizational structure.

\subsection{Skill-Guided Spatial Reasoning}

During inference episode $t$, \method performs state-adaptive spatial reasoning over $(I_t,q_t)$. As shown in the left panel of Figure~\ref{fig:framework}, each iteration follows four stages: Skill Retrieval, Action Selection, Tool/Code Execution, and State Update. Figure~\ref{fig:example} illustrates how these skills organize tool execution and are subsequently refined through failure diagnosis. With $\mathrm{Lib}_t$ fixed, the loop repeats until the accumulated state is sufficient to produce an answer.

\subsubsection{Skill Retrieval}

At each iteration, a retrieval interface converts the question and current state into a structured decision context and queries the skill library for a small candidate set:
\begin{equation}
\mathcal{C}_{t,k}
=\mathrm{LLM}_{\mathrm{Retrieve}}(I_t,q_t,M_{t,k};\mathcal{S}_t).
\end{equation}
Here $\mathcal{C}_{t,k}\subseteq\mathcal{S}_t$ may contain Tool-Use or Geometry Skills. The interface is implemented with task-specific instruction templates that expose the target and reference objects, relation, available evidence, and required geometric operation.

\subsubsection{Action Selection}

Given the retrieved candidates and current state, the controller selects the next action:
\begin{equation}
\begin{gathered}
u_{t,k}
=\mathrm{LLM}_{\mathrm{Select}}(I_t,q_t,M_{t,k},\mathcal{C}_{t,k}),\\[-2pt]
u_{t,k}
\in\{\mathrm{stop}\}\cup
\bigl(\mathcal{C}_{t,k}\cap\mathcal{S}_{\mathrm{tool},t}\bigr)
\cup\mathcal{A}_{\mathrm{tool},t}.
\end{gathered}
\end{equation}
The controller may terminate with the current answer, activate or resume a retrieved Tool-Use Skill, or directly call a compatible tool atom. Geometry Skills are not independent controller actions; they are instantiated within the Python atom during execution.

\input{Tables/subtask_acc_breakdown}

\subsubsection{Tool/Code Execution}

The selected action executes either a single tool atom or the next dependency-ready node of an active Tool-Use Skill. The resulting observation is returned immediately for state update, so the skill can be interrupted after any node. Visual tool atoms return perceptual evidence directly. When execution reaches the Python atom, the agent binds the required artifacts from $M_{t,k}$ and instantiates a compatible retrieved Geometry Skill. Its code blocks execute in dependency order and return both numerical outputs and verifier outcomes to the Python node. This preserves a compact tool-level action space while making geometric computation structured and verifiable.

\subsubsection{State Update}

After execution, the controller incorporates the new observation and verifier outcomes into its working memory. For a non-terminal action, the complete iteration is
\begin{equation}
\begin{aligned}
\bigl(I_t,q_t,M_{t,k}\bigr)
&\xrightarrow{\mathrm{Retrieve}}\mathcal{C}_{t,k}
\xrightarrow{\mathrm{Select}}u_{t,k},\\[-2pt]
u_{t,k}
&\xrightarrow{\mathrm{Exec}}(o_{t,k},\nu_{t,k})
\xrightarrow{\mathrm{Update}}M_{t,k+1}.
\end{aligned}
\end{equation}
Here $o_{t,k}$ is the execution output, while $\nu_{t,k}\in\{0,1\}$ indicates whether execution and local verification succeed. Unless the controller selects $\mathrm{stop}$, the updated state initiates the next retrieval round; otherwise, the current answer hypothesis is returned as the final answer. After final-answer evaluation, the completed trace is stored for later batch-wise self-evolution.

\subsection{Skill Self-Evolution}

\method evolves a single skill library from batches of completed trajectories. The trajectory produced by episode $t$ is
\begin{equation}
\begin{aligned}
\tau_t
&=\left(
\bigl((M_{t,k},u_{t,k},o_{t,k},\nu_{t,k})\bigr)_{k=1}^{K_t},
\hat{y}_t
\right),\\
\overline{\mathcal{B}}_{t+1}
&=\mathcal{B}_{t}\cup\{\tau_t\}.
\end{aligned}
\end{equation}
where $K_t$ is the number of executed node-level steps, $\hat{y}_t$ is the final answer, and $\overline{\mathcal{B}}_{t+1}$ is the buffer before evolution. Self-evolution is triggered only when $|\overline{\mathcal{B}}_{t+1}|\geq N_{\mathrm{buf}}$; otherwise, $\mathrm{Lib}_{t+1}=\mathrm{Lib}_t$ and $\mathcal{B}_{t+1}=\overline{\mathcal{B}}_{t+1}$. Once triggered, \method analyzes the buffered trajectories, fuses the extracted structures with the current library, and prunes unreliable or inactive entries. The processed buffer is then cleared, while its aggregated invocation statistics are retained.

\input{Tables/RQ2_tool}

\subsubsection{Trajectory Analysis}

A single trajectory may contain incidental failures or task-specific choices, so \method extracts reusable structures only after accumulating sufficient evidence in the buffer. Each buffered trajectory is decomposed into executed operations, arguments, produced artifacts, consumed variables, and local outcomes. The analyzer then aggregates their dependency patterns and returns atomic-instruction updates together with supported and invalid skill fragments:
\begin{equation}
\bigl(
\Delta\mathcal{A}_{t},
\widehat{\mathcal{G}}^{+}_{t},
\widehat{\mathcal{G}}^{-}_{t}
\bigr)
=\mathrm{LLM}_{\mathrm{Analyze}}(\overline{\mathcal{B}}_{t+1}).
\end{equation}
Here $\Delta\mathcal{A}_{t}$ contains newly required or revised atomic instructions, while $\widehat{\mathcal{G}}^{+}_{t}$ and $\widehat{\mathcal{G}}^{-}_{t}$ contain supported and invalid skill fragments, respectively. For example, a supported fragment may require invoking atom $a_i$ before $a_j$, whereas an invalid fragment records that this transition repeatedly causes execution or verifier failures. Aggregating multiple trajectories reduces the influence of isolated errors and exposes dependencies that recur across tasks.

\subsubsection{Skill Fusion}

Extracted atoms and fragments may duplicate, extend, or conflict with existing skills. The fusion module therefore compares them with the unified library and produces a structurally revised library:
\begin{equation}
\mathrm{Lib}^{\mathrm{fuse}}_{t+1}
=\mathrm{LLM}_{\mathrm{Fusion}}\bigl(
\mathrm{Lib}_{t},\Delta\mathcal{A}_{t},
\widehat{\mathcal{G}}^{+}_{t},
\widehat{\mathcal{G}}^{-}_{t}\bigr).
\end{equation}
We write $\mathrm{Lib}^{\mathrm{fuse}}_{t+1}=(\mathcal{A}^{\mathrm{fuse}}_{t+1},\mathcal{S}^{\mathrm{fuse}}_{t+1})$. Fusion first reconciles $\Delta\mathcal{A}_{t}$ with $\mathcal{A}_{t}$ by adding a novel typed instruction or replacing an existing implementation when the new version better satisfies its local verifier. It then edits high-level skills through three operations: adding a supported fragment, deleting a redundant or incompatible dependency, or replacing an affected subgraph while preserving the remainder of the skill. Tool-organizing fragments update Tool-Use Skills, whereas fragments that transform or verify numerical evidence update Geometry Skills. All edits are applied directly to the unified library.

\subsubsection{Skill Pruning}

Fusion revises the library structure, while pruning removes unreliable or inactive entries according to their invocation history. For each $z\in\mathcal{A}^{\mathrm{fuse}}_{t+1}\cup\mathcal{S}^{\mathrm{fuse}}_{t+1}$, let $n_t(z)$ be its number of invocations and $\mathrm{ER}_t(z)$ the fraction of invocations that end in an execution or local-verifier failure. We further define its inactivity age as $\mathrm{Age}_t(z)=t-t_{\mathrm{last}}(z)$ and its tenure as $\mathrm{Tenure}_t(z)=t-t_{\mathrm{add}}(z)$, where $t_{\mathrm{last}}(z)$ and $t_{\mathrm{add}}(z)$ denote the most recent invocation and insertion episodes, respectively. Before the first invocation, we set $t_{\mathrm{last}}(z)=t_{\mathrm{add}}(z)$.

Given an error-rate threshold $\rho_{\mathrm{err}}$, an inactivity limit $T_{\mathrm{idle}}$, and a minimum support threshold $n_{\min}$, we prune $z$ when
\begin{equation}
\begin{aligned}
\mathrm{Prune}_t(z)
={}&
\bigl[n_t(z)\geq n_{\min}
\wedge \mathrm{ER}_t(z)>w_t(z)\rho_{\mathrm{err}}\bigr]\\[-2pt]
&{}\vee
\bigl[\mathrm{Age}_t(z)>w_t(z)T_{\mathrm{idle}}\bigr].
\end{aligned}
\label{eq:skill_pruning}
\end{equation}
Both thresholds are adapted by the tenure-dependent retention factor
\[
w_t(z)=1+\lambda
\frac{\mathrm{Tenure}_t(z)}
{1+\mathrm{Tenure}_t(z)},
\]
where $\lambda\geq0$ is chosen such that $(1+\lambda)\rho_{\mathrm{err}}\leq1$. Thus, $w_t(z)$ grows from $1$ toward $1+\lambda$, providing longer-lived entries with bounded protection, while $n_{\min}$ prevents pruning after only a few noisy failures. Entries without any invocation history are evaluated only by inactivity. After removing all marked atoms and skills, we repair or discard skills that depend on a pruned atom, yielding the updated library $\mathrm{Lib}_{t+1}$.

%% file: Figures/skill_examples.tex
\begin{figure}[t]
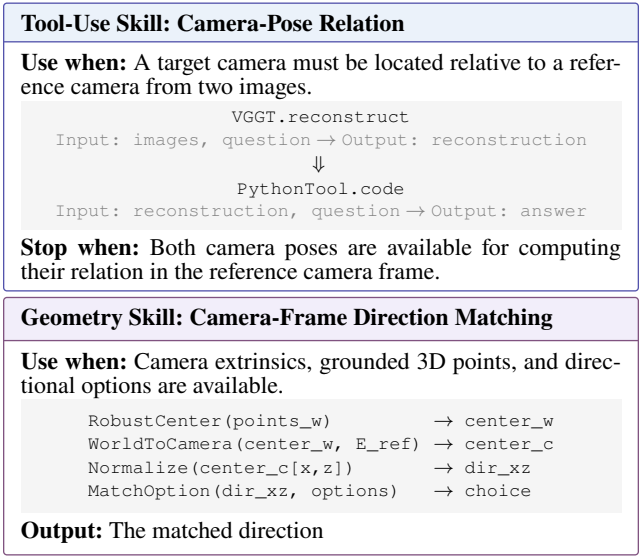

    \centering
    \small
    \definecolor{toolheader}{RGB}{235,242,250}
    \definecolor{geoheader}{RGB}{242,238,250}
    \definecolor{codebg}{gray}{0.96}
    \definecolor{keyword}{RGB}{0,0,180}
    \definecolor{comment}{RGB}{0,120,0}
    \definecolor{string}{RGB}{163,21,21}

    \begin{tcolorbox}[
        colback=white,
        colframe=blue!35!gray,
        boxrule=0.5pt,
        arc=1pt,
        left=1.2mm,
        right=1.2mm,
        top=0.45mm,
        bottom=0.45mm,
        toptitle=0.35mm,
        bottomtitle=0.35mm,
        before skip=0pt,
        after skip=0.7mm,
        title=\textbf{Tool-Use Skill: Camera-Pose Relation},
        colbacktitle=toolheader,
        coltitle=black,
        fonttitle=\bfseries\small,
        fontupper=\footnotesize\linespread{0.90}\selectfont
    ]

    \textbf{Use when:}
    A target camera must be located relative to a reference camera
    from two images.

    \vspace{0.25mm}

    \begin{tcolorbox}[
        colback=codebg,
        boxrule=0pt,
        frame hidden,
        arc=0pt,
        left=0.8mm,
        right=0.8mm,
        top=0.35mm,
        bottom=0.35mm,
        before skip=0pt,
        after skip=0pt
    ]
    \centering
    \scriptsize\linespread{0.88}\selectfont

    \texttt{VGGT.reconstruct}\\[-0.1mm]
    {\color{gray}
    \texttt{Input: images, question}
     $\rightarrow$
    \texttt{Output: reconstruction}}

    \vspace{0.15mm}
    $\Downarrow$
    \vspace{0.15mm}

    \texttt{PythonTool.code}\\[-0.1mm]
    {\color{gray}
    \texttt{Input: reconstruction, question}
     $\rightarrow$
    \texttt{Output: answer}}

    \end{tcolorbox}

    \vspace{1.2mm}

    \textbf{Stop when:}
    Both camera poses are available for computing their relation
    in the reference camera frame.

    \end{tcolorbox}

    \begin{tcolorbox}[
        colback=white,
        colframe=violet!35!gray,
        boxrule=0.5pt,
        arc=1pt,
        left=1.2mm,
        right=1.2mm,
        top=0.45mm,
        bottom=0.45mm,
        toptitle=0.35mm,
        bottomtitle=0.35mm,
        before skip=0pt,
        title=\textbf{Geometry Skill: Camera-Frame Direction Matching},
        colbacktitle=geoheader,
        coltitle=black,
        fonttitle=\bfseries\small,
        fontupper=\footnotesize\linespread{0.90}\selectfont
    ]

    \textbf{Use when:}
    Camera extrinsics, grounded 3D points, and directional options
    are available.

    \vspace{0.25mm}

    \begin{tcolorbox}[
        colback=codebg,
        boxrule=0pt,
        frame hidden,
        arc=0pt,
        left=0.8mm,
        right=0.8mm,
        top=0.3mm,
        bottom=0.3mm,
        before skip=0pt,
        after skip=0pt
    ]
    \centering
    \scriptsize\linespread{0.88}\selectfont
    \renewcommand{\arraystretch}{1.0}
    \begin{tabular}{@{}l@{\;\;}c@{\;\;}l@{}}
    \texttt{RobustCenter(points\_w)}
    & $\rightarrow$ &
    \texttt{center\_w}
    \\
    \texttt{WorldToCamera(center\_w, E\_ref)}
    & $\rightarrow$ &
    \texttt{center\_c}
    \\
    \texttt{Normalize(center\_c[x,z])}
    & $\rightarrow$ &
    \texttt{dir\_xz}
    \\
    \texttt{MatchOption(dir\_xz, options)}
    & $\rightarrow$ &
    \texttt{choice}

    \end{tabular}
    \end{tcolorbox}

    \vspace{1.2mm}

    \textbf{Output:}
    The matched direction

    \end{tcolorbox}

    \caption{Examples of Tool-Use and Geometry Skills.}
    \label{fig:skill_examples}
    \vspace{-3mm}
\end{figure}

%% file: Tables/subtask_acc_breakdown.tex
\begin{table*}[t]
    \centering
    \caption{Average accuracy (\%) over the online stream, broken down by dataset-specific subtasks and capability groups. The best and second-best results within each backbone block are highlighted in bold and underlined, respectively.}
    \label{tab:subtask_acc_breakdown}
    \setlength{\tabcolsep}{3pt}
    \resizebox{0.95\textwidth}{!}{%
    \begin{tabular}{lcccc>{\columncolor{black!5}}cccc>{\columncolor{black!5}}ccc>{\columncolor{black!5}}c}
        \toprule
        Method & \multicolumn{5}{c}{MMSI} &
        \multicolumn{4}{c}{MindCube} &
        \multicolumn{3}{c}{OmniSpatial} \\
        \cmidrule(lr){2-6}\cmidrule(lr){7-10}\cmidrule(lr){11-13}
        &
        Attr. & Motion & Pos. Rel. & MSR & Overall &
        Rot. & Ard. & Amg. & Overall &
        Dyn. & Persp. & Overall \\
        \midrule
        \rowcolor{blue!10}[\tabcolsep][\tabcolsep]
        \multicolumn{13}{l}{\textbf{GPT-5.4}} \\
        SpaAge-PE
        & 38.24 & 48.00 & 37.50 & 40.00 & 39.50
        & 57.14 & \textbf{64.71} & 57.99 & 58.50
        & 55.42 & 53.85 & 54.50 \\
        SpaAge-ReAct
        & \textbf{52.94} & 52.00 & 39.58 & 42.22 & 44.00
        & \textbf{71.43} & 58.82 & 52.66 & 54.50
        & 59.04 & 53.85 & 56.00 \\
        LAST
        & 37.64 & 51.00 & 42.37 & \underline{48.79} & 44.09
        & 48.92 & \underline{60.23} & \underline{68.24} & 66.21
        & 60.24 & \underline{55.63} & 57.54 \\
        GCA
        & 35.29 & 52.00 & \underline{48.96} & \textbf{53.33} & \underline{48.00}
        & \underline{64.29} & 52.94 & \textbf{73.37} & \underline{71.01}
        & 63.89 & \textbf{58.97} & \textbf{61.00} \\
        \noalign{\vskip\aboverulesep}
        \hdashline
        \noalign{\vskip\belowrulesep}
        AWM
        & 31.25 & \underline{54.24} & 45.64 & 40.00 & 43.01
        & 53.85 & 38.46 & 57.93 & 56.14
        & \underline{65.43} & 52.88 & 58.38 \\
        ASI
        & 38.71 & 52.17 & 31.18 & 42.50 & 37.43
        & 42.86 & 41.18 & 50.89 & 49.50
        & 63.86 & 52.99 & 57.50 \\
        \midrule
        \method
        & \underline{41.18} & \textbf{56.00} & \textbf{52.08} & 46.67 & \textbf{49.50}
        & \underline{64.29} & 58.82 & \textbf{73.37} & \textbf{71.50}
        & \textbf{66.27} & 55.56 & \underline{60.27} \\
        \midrule
        \rowcolor{purple!10}[\tabcolsep][\tabcolsep]
        \multicolumn{13}{l}{\textbf{Gemini 2.5 Pro}} \\
        SpaAge-PE
        & 41.21 & 47.96 & 52.13 & \textbf{57.82} & 51.00
        & 71.52 & 58.91 & 69.79 & 69.08
        & 51.82 & 59.81 & 56.37 \\
        SpaAge-ReAct
        & 44.11 & 48.00 & 57.29 & 51.13 & 52.50
        & 64.19 & 58.83 & \underline{74.50} & \underline{72.38}
        & 59.02 & 55.53 & 57.08 \\
        LAST
        & 30.67 & 48.00 & 58.23 & 54.28 & 51.38
        & 68.04 & \underline{70.78} & 60.14 & 61.60
        & 60.25 & 53.64 & 56.38 \\
        GCA
        & 32.35 & 52.00 & 60.42 & 48.89 & 52.00
        & 85.67 & \textbf{76.47} & 62.13 & 65.03
        & 61.45 & 61.54 & 61.53 \\
        \noalign{\vskip\aboverulesep}
        \hdashline
        \noalign{\vskip\belowrulesep}
        AWM
        & \textbf{55.88} & 60.00 & \underline{65.62} & \underline{55.56} & \underline{61.00}
        & 78.57 & 64.71 & 61.54 & 63.00
        & \underline{65.06} & \underline{68.38} & \underline{66.97} \\
        ASI
        & 50.00 & \underline{68.00} & 64.58 & 51.11 & 59.70
        & \underline{85.71} & 70.58 & 59.76 & 62.50
        & 63.86 & 64.10 & 64.01 \\
        \midrule
        \method
        & \underline{52.94} & \textbf{72.03} & \textbf{66.67} & 53.33 & \textbf{61.50}
        & \textbf{85.73} & 64.71 & \textbf{84.02} & \textbf{82.50}
        & \textbf{67.47} & \textbf{73.50} & \textbf{70.94} \\
        \bottomrule
    \end{tabular}
    }
\end{table*}

%% file: Tables/RQ2_tool.tex
\begin{figure*}[t]
\centering
\small
\setlength{\tabcolsep}{3pt}
\newsavebox{\rqtwotablebox}
\newlength{\rqtwotableheight}
\sbox{\rqtwotablebox}{%
\resizebox{0.71\textwidth}{!}{%
\begin{tabular}{lccc>{\columncolor{black!5}}cccc>{\columncolor{black!5}}c}
\toprule
& \multicolumn{4}{c}{FTCS $\uparrow$}
& \multicolumn{4}{c}{Python Success Rate $\uparrow$} \\
\cmidrule(lr){2-5}
\cmidrule(lr){6-9}
Method
& MMSI & MindCube & OmniSpatial & Avg.
& MMSI & MindCube & OmniSpatial & Avg. \\
\midrule
\rowcolor{blue!10}[\tabcolsep][\tabcolsep]
\multicolumn{9}{l}{\textbf{GPT-5.4}} \\
SpaAge
& 77.08 & 87.50 & \underline{90.39} & 84.99 & \underline{94.45} & 85.71 & 86.19 & \underline{88.78} \\

LAST
& 83.32 & 90.00 & 87.64 & 86.99 & 92.27 & 80.32 & 65.23 & 79.27 \\

GCA
& 85.35 & 90.71 & 83.73 & 86.60 & 90.99 & \textbf{90.38} & 79.76 & 87.04 \\

AWM
& \underline{88.75} & \textbf{91.84} & 81.58 & \underline{87.39} & 79.01 & 75.41 & 57.63 & 70.68 \\

ASI
& 83.33 & 85.19 & 85.71 & 84.74 & 77.78 & 83.33 & \underline{92.68} & 84.60 \\

\midrule
\textsc{NeSy-Spatial}
& \textbf{97.89} & \underline{91.19} & \textbf{94.59} & \textbf{94.56} & \textbf{97.76} & \underline{89.20} & \textbf{99.39} & \textbf{95.45} \\
\midrule
\rowcolor{purple!10}[\tabcolsep][\tabcolsep]
\multicolumn{9}{l}{\textbf{Gemini 2.5 Pro}} \\
SpaAge
& 33.53 & 26.76 & 54.01 & 38.10 & 35.05 & 14.64 & 64.66 & 38.12 \\

LAST
& \textbf{85.18} & 82.17 & 83.06 & 83.47 & 69.75 & 82.31 & 78.23 & 76.76 \\

GCA
& 65.15 & 85.14 & 87.56 & 79.28 & 70.25 & \underline{91.86} & 80.56 & 80.89 \\

AWM
& \underline{84.53} & \underline{91.43} & \underline{89.44} & \underline{88.46} & 91.07 & 81.63 & \underline{92.65} & \underline{88.45} \\

ASI
& 79.87 & 80.00 & 77.98 & 79.28 & \underline{95.22} & 56.03 & 78.85 & 76.70 \\

\midrule
\textsc{NeSy-Spatial}
& 81.71 & \textbf{92.96} & \textbf{91.33} & \textbf{88.67} & \textbf{97.45} & \textbf{97.86} & \textbf{93.33} & \textbf{96.21} \\
\bottomrule
\end{tabular}
}
}
\setlength{\rqtwotableheight}{%
\dimexpr\ht\rqtwotablebox+\dp\rqtwotablebox\relax}
\begin{minipage}[t]{0.715\textwidth}%
\vspace{0pt}
\noindent\usebox{\rqtwotablebox}%
\end{minipage}\hspace{0.005\textwidth}%
\begin{minipage}[t]{0.28\textwidth}%
\vspace{0pt}
\centering
\includegraphics[
height=1.05\rqtwotableheight
]{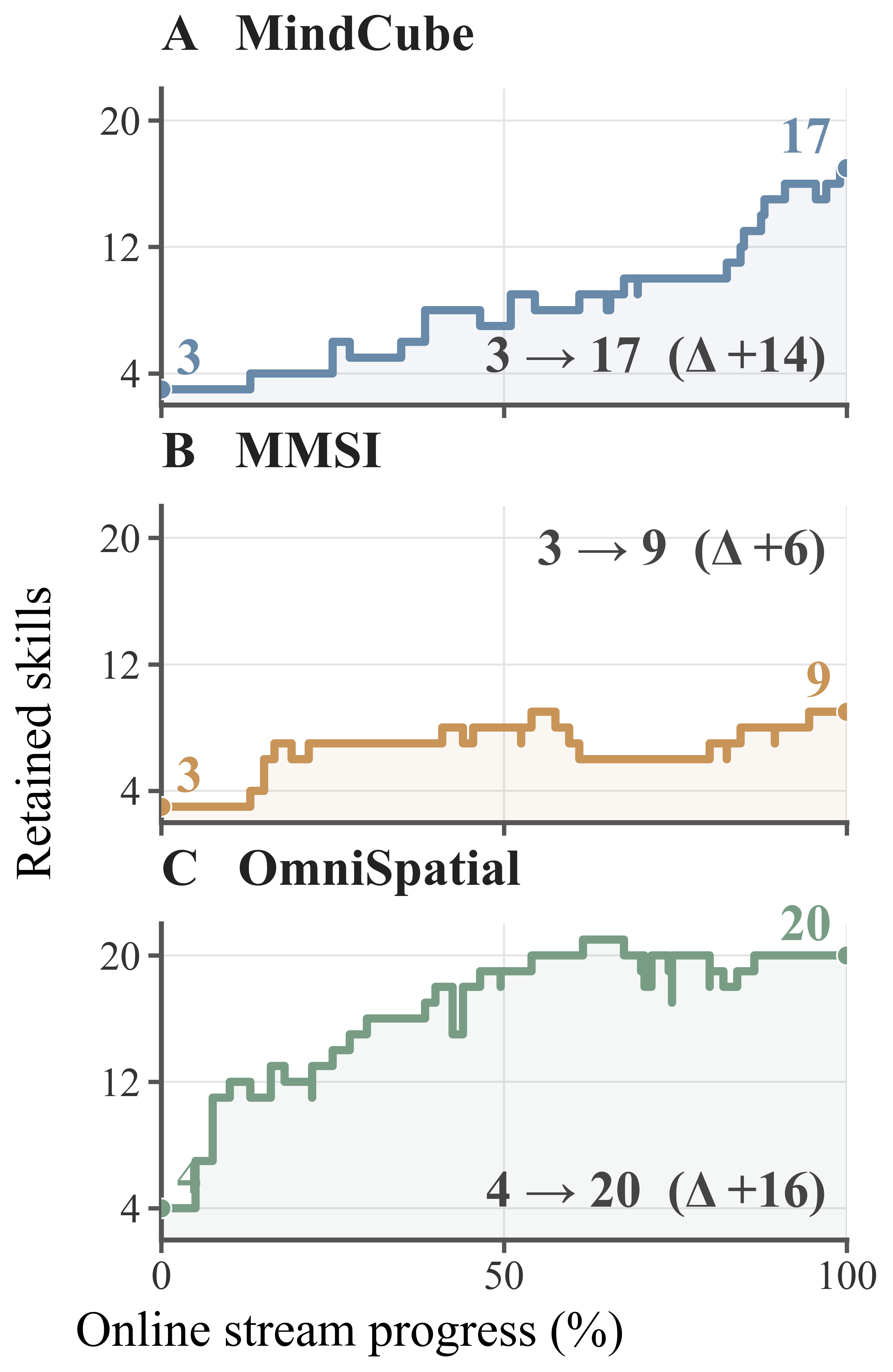}
\end{minipage}
\caption{
\textbf{Left:} FTCS and Python success rate across spatial reasoning benchmarks; the best and second-best results within each block are highlighted in bold and underlined, respectively.
\textbf{Right:} the evolution of the number of retained skills throughout the online stream.
}
\label{fig:tool_reliability_and_skill_evolution}
\end{figure*}

%% file: 04_experiments.tex
\section{Experiments}


\subsection{Experimental Setup}

\noindent\textbf{Datasets and Models.}
We evaluate \method on MMSI~\cite{yang2025mmsi}, MindCube~\cite{chen2026mindcube}, and OmniSpatial~\cite{jia2025omnispatial} using GPT-5.4 and Gemini 2.5 Pro as the two API backbones. For MMSI, we use all four subsets: Attribute, Motion, Positional Relation, and Multi-Step Reasoning (MSR). For MindCube, we use all three subsets: Rotation, Around, and Among. Following GCA~\cite{chen2026gca}, we evaluate OmniSpatial on two subsets, Dynamic Reasoning and Perspective Taking.

\noindent\textbf{Comparison Methods.}
We compare against the general inductive agents AWM~\cite{wang2024agent} and ASI~\cite{wang2025inducing}, as well as domain-specific spatial reasoning agents with different tool-use mechanisms. SpaAge-PE and SpaAge-ReAct from SpatialScore~\cite{wu2026spatialscore} let the LVLM plan tool calls directly over a provided toolbox, whereas GCA~\cite{chen2026gca} first maps each problem into a unified geometric formalization before invoking tools. All methods are evaluated with the same tool set and interface configuration; LAST~\cite{tian2026last} differs in that its spatial skills are predefined.
For all methods evaluated in the online setting, we report the average accuracy over all samples in the online stream.

\noindent\textbf{Evaluation Metrics.}
We report prequential answer accuracy, where each prediction is evaluated before its label is used for skill evolution. Tool execution is measured by First-Pass Compositional Tool-Chain Success Rate (FTCS) and Python success rate. For samples $\mathcal{S}_{\mathrm{pipe}}$ invoking at least two distinct task tools,
\[
\mathrm{FTCS}
=
\frac{1}{|\mathcal{S}_{\mathrm{pipe}}|}
\sum_{s\in\mathcal{S}_{\mathrm{pipe}}} c_s,
\]
where $c_s=1$ if all task-tool calls succeed without a failed attempt, and $0$ otherwise. Python success rate is the fraction of successful \texttt{PythonTool.code} calls, counting retries separately.

\noindent\textbf{Implementation Details.}
We initialize the skill library with $N_{\mathrm{seed}}=3$ seed skills for MMSI and MindCube, and $N_{\mathrm{seed}}=4$ for OmniSpatial. Skill evolution is triggered after every $N_{\mathrm{buf}}=20$ completed trajectories. For skill pruning, we set the minimum invocation support to $n_{\min}=5$, the base error-rate threshold to $\rho_{\mathrm{err}}=0.4$, the inactivity limit to $T_{\mathrm{idle}}=100$ episodes, and the tenure-dependent retention coefficient to $\lambda=0.5$, satisfying $(1+\lambda)\rho_{\mathrm{err}}\leq 1$. The same hyperparameters are used across all datasets and LVLM backbones. We provide all prompts used in the paper, the complete set of evolved rules, and the remaining hyperparameter settings in the appendix.

\subsection{Main Results}

\textbf{\method improves spatial reasoning performance and achieves state-of-the-art results.} As shown in Table~\ref{tab:subtask_acc_breakdown}, \method achieves the best Overall accuracy in five of the six backbone--dataset settings. With GPT-5.4, \method reaches 49.50 on MMSI and 71.50 on MindCube, outperforming the strongest baselines by 1.50 and 0.49 percentage points, respectively, while remaining competitive on OmniSpatial (60.27 vs.\ 61.00). With Gemini 2.5 Pro, it achieves the highest Overall accuracy on all three benchmarks, outperforming the strongest baselines by 0.50, 10.12, and 3.97 percentage points on MMSI, MindCube, and OmniSpatial, respectively. These consistent gains across different backbones and benchmarks demonstrate the effectiveness of \method for spatial reasoning.

\textbf{\method improves tool utilization efficiency and execution accuracy.} The left panel of Figure~\ref{fig:tool_reliability_and_skill_evolution} reports SpaAge-PE and SpaAge-ReAct jointly as SpaAge. With GPT-5.4, \method achieves the highest average FTCS of 94.56 and Python success rate of 95.45, outperforming the strongest baseline averages by 7.17 and 6.67 percentage points, respectively. Across datasets, it leads both metrics on MMSI and OmniSpatial and remains close to the best on MindCube, demonstrating consistent gains beyond the averages. With Gemini 2.5 Pro, \method again achieves the best average FTCS and Python success rate, exceeding the strongest baselines by 0.21 and 7.76 points, respectively. It attains the highest Python success rate on all three benchmarks and the highest FTCS on MindCube and OmniSpatial, while trailing LAST by 3.47 points in FTCS on MMSI. Together, these results show that \method more reliably completes compositional tool chains on the first pass while improving the accuracy of individual Python executions.


\subsection{Further Analysis}

\begin{figure}[t]
    \centering
    \includegraphics[width=\columnwidth]{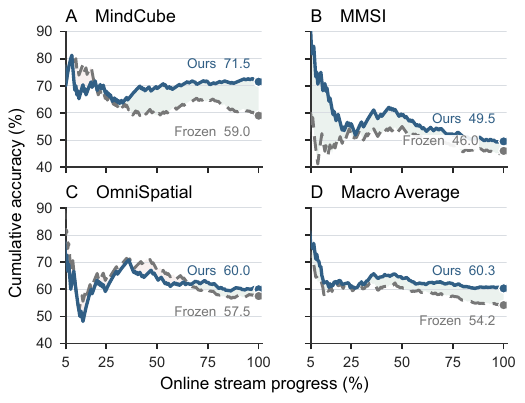}
    \caption{Cumulative prequential accuracy of \method and Frozen throughout the online stream. Frozen keeps the initial skill library fixed without online evolution.}
    \label{fig:rq3_skill_evolution}
\end{figure}

\textbf{Online skill evolution yields consistent overall gains in spatial reasoning performance.} Figure~\ref{fig:rq3_skill_evolution} compares the cumulative prequential accuracy of \method with Frozen, whose initial skill library (i.e., the knowledge base) remains fixed throughout the stream, corresponding to the setting without evolution. Because cumulative estimates based on only a few observations are highly volatile, the curves are displayed from 5\% stream progress, after 10 samples have been processed; each point nevertheless includes all observations from the start of the stream. The curves become more stable as evaluation proceeds, and \method establishes sustained gains over Frozen across all three benchmarks. By the end of the stream, \method finishes 12.5, 3.5, and 2.5 percentage points higher on MindCube, MMSI, and OmniSpatial, respectively. The macro-average accuracy similarly improves from 54.2\% to 60.3\%, showing that online skill evolution provides consistent benefits across the three benchmarks. Together, the endpoint margins and full-stream trajectories show that these gains persist as evaluation continues rather than arising from a brief early advantage.

\textbf{Skill evolution selectively expands the library rather than accumulating skills indiscriminately.} The right panel of Figure~\ref{fig:tool_reliability_and_skill_evolution} tracks how the number of retained skills changes throughout online evolution. Overall, the library expands from 3 to 17 skills on MindCube, from 3 to 9 on MMSI, and from 4 to 20 on OmniSpatial. However, this growth is not monotonic: the library size decreases at multiple stages as unreliable or inactive skills are removed. These downward steps indicate that the pruning strategy actively controls redundancy and prevents unbounded accumulation. The resulting trajectories reflect a dynamic balance between incorporating reusable skills from new experience and retaining only those that remain useful.


\input{Tables/ablation}

\textbf{Each component plays a distinct and complementary role in the full framework.} As shown in Table~\ref{tab:ablation_pipeline_geometry}, disabling skill evolution while retaining both skill types reduces accuracy by 12.50, 3.50, and 2.78 percentage points on MindCube, MMSI, and OmniSpatial, respectively. Removing Tool-Use Skills causes corresponding drops of 11.50, 6.00, and 2.68 points, whereas removing Geometry Skills lowers accuracy by 2.00, 2.50, and 2.27 points. The full configuration consistently performs best across all three benchmarks, demonstrating that online evolution provides substantial gains while Tool-Use and Geometry Skills contribute complementary execution and reasoning capabilities.


%% file: Tables/ablation.tex
\begin{table}[t]
\centering
\small
\caption{Ablation study of skill evolution, Tool-Use Skills, and Geometry Skills using GPT-5.4.}
\label{tab:ablation_pipeline_geometry}
\setlength{\tabcolsep}{3pt}
\resizebox{\columnwidth}{!}{%
\begin{tikzpicture}[baseline=(ablationtab.base)]
\node[inner sep=0pt, outer sep=0pt] (ablationtab) {%
\begin{tabular}{cccccc}
\toprule
\multicolumn{2}{c}{Skill Type}
& \multirow{2}{*}{Evolution}
& \multicolumn{3}{c}{Accuracy (\%) $\uparrow$} \\
\cmidrule(lr){1-2}
\cmidrule(lr){4-6}
Tool-Use
& Geometry
&
& MindCube
& MMSI
& OmniSpatial \\
\midrule
$\checkmark$ & $\checkmark$ & & 59.00 & 46.00 & 57.49 \\
& $\checkmark$ & $\checkmark$ & 60.00 & 43.50 & 57.59 \\
$\checkmark$ & & $\checkmark$ & 69.50 & 47.00 & 58.00 \\
$\checkmark$ & $\checkmark$ & $\checkmark$ & 71.50 & 49.50 & 60.27\\
\bottomrule
\end{tabular}
};
\draw[dash pattern=on 2pt off 1.5pt, line width=0.4pt]
    ($(ablationtab.north west)!0.482!(ablationtab.north east)$)
    --
    ($(ablationtab.south west)!0.482!(ablationtab.south east)$);
\end{tikzpicture}%
}
\end{table}

%% file: 05_conclusion.tex
\section{Conclusion}

This paper presented \method, a self-evolving neuro-symbolic framework for tool-augmented spatial reasoning. It organizes typed atomic instructions for tool interaction and geometric computation into interruptible Tool-Use Skills and reusable Geometry Skills. Across three benchmarks and two LVLM backbones, \method achieves the best aggregate accuracy in five of six settings and the highest average FTCS and Python success rate. Online results show sustained accuracy gains with experience, while library-size changes indicate that pruning removes unreliable or inactive skills instead of allowing indiscriminate growth. Evaluation is limited to selected benchmarks and tool environments; future work will examine broader tasks and transfer to unseen settings.

%% file: Appendix/appendix.tex
\clearpage
\appendix

\section*{Supplementary Material}

Section~\ref{app:implementation} reports implementation settings, native
tools, and skill representations. Section~\ref{app:case_studies} presents
end-to-end inference and online-evolution cases. Section~\ref{app:evolved-skill-inventory}
summarizes the retained learned skills, and Section~\ref{app:prompt_templates}
provides the complete prompt templates used for inference and evolution.
\medskip

\input{Appendix/A.tex}

\input{Appendix/B.tex}
\input{Appendix/C.tex}
\input{Appendix/D.tex}

%% file: Appendix/A.tex
\section{Implementation Details}
\label{app:implementation}

\subsection{Hyperparameter Settings}
\label{app:hyperparameters}

We report the implementation settings used for skill initialization,
online induction, candidate fusion, empirical skill maintenance, and
inference. Most settings are shared across datasets and LVLM backbones,
while dataset-specific settings are listed separately.

\noindent\textbf{Skill Initialization.}
We initialize the online library with a small set of dataset-specific
Tool-Use seed skills. These seeds provide initial pipeline structures,
while the mutable skill library is subsequently updated from online
interaction trajectories.

\smallskip
\noindent\textbf{Online Induction and Fusion.}
Online induction starts after an initial warm-up period. Interaction
traces are organized by routing family before being summarized into
candidate skills. Candidate skills are subsequently fused with the
mutable skill library.

\smallskip
\noindent\textbf{Skill Selection and Library Capacity.}
At inference time, the selector first observes compact descriptions of
the available skills and retrieves a small subset relevant to the
current problem. Complete executable definitions are exposed only for
the selected skills, reducing prompt length and irrelevant context.

\smallskip
\noindent\textbf{Empirical Skill Maintenance.}
Newly induced skills enter a probationary state. Their invocation
statistics and task outcomes are accumulated during subsequent
episodes. A skill can be promoted after obtaining sufficient empirical
support, whereas consistently unsuccessful skills are removed from the
mutable library. The initial seed skills are protected from this
retirement procedure.

\begin{table}[H]
    \centering
    \footnotesize
    \setlength{\tabcolsep}{4pt}
    \renewcommand{\arraystretch}{1.06}
    \caption{Shared settings for initialization, online induction,
    selection, and empirical maintenance.}
    \label{tab:appendix_shared_settings}
    \begin{tabularx}{\columnwidth}{
        @{}
        >{\raggedright\arraybackslash}X
        >{\centering\arraybackslash}p{0.15\columnwidth}
        @{}
    }
        \toprule
        \rowcolor{cardtitle}
        \textbf{Setting} & \textbf{Value} \\
        \midrule
        \rowcolor{cardsubtle}
        \multicolumn{2}{@{}l}{\textbf{Initialization}} \\
        MMSI seed skills & 3 \\
        MindCube seed skills & 3 \\
        OmniSpatial seed skills & 4 \\
        \midrule
        \rowcolor{cardsubtle}
        \multicolumn{2}{@{}l}{\textbf{Online induction and fusion}} \\
        Warm-up trajectories & 10 \\
        Same-family trace batch & 5 \\
        Repeated-failure support & 2 \\
        Pending-trace fallback threshold & 40 \\
        Candidate fusion batch & 2 \\
        Maximum traces per induction prompt & 20 \\
        Prototype trace window & 80 \\
        \midrule
        \rowcolor{cardsubtle}
        \multicolumn{2}{@{}l}{\textbf{Skill selection and capacity}} \\
        Maximum selected skills per query & 4 \\
        Maximum skills in the mutable library & 24 \\
        \midrule
        \rowcolor{cardsubtle}
        \multicolumn{2}{@{}l}{\textbf{Empirical maintenance}} \\
        Zero-correct retirement support & 2 \\
        Minimum invocation support & 5 \\
        Minimum retained accuracy & 0.4 \\
        Inactivity threshold & 100 \\
        \bottomrule
    \end{tabularx}
\end{table}

\noindent\textbf{Runtime and Inference Settings.}
All experiments use the simple agent runtime. Queries are processed in
waves of five. We allow a larger reasoning budget for MMSI because its
questions typically require longer multi-view geometric pipelines.
GroundingDINO is fixed as the object-detection backend across LVLM
backbones, preventing changes in the reasoning model from simultaneously
changing the visual grounding component.

\begin{table}[H]
    \centering
    \small
    \setlength{\tabcolsep}{4pt}
    \caption{Dataset-specific inference settings.}
    \label{tab:appendix_inference}
    \begin{tabularx}{\columnwidth}{lYY}
        \toprule
        \rowcolor{cardtitle}
        \textbf{Dataset} & \textbf{Maximum turns} & \textbf{Wave size} \\
        \midrule
        MMSI        & 12 & 5 \\
        MindCube    & 8  & 5 \\
        OmniSpatial & 8  & 5 \\
        \bottomrule
    \end{tabularx}
\end{table}

\subsection{Native Tool Library}
\label{app:native_tools}

The agent is equipped with a fixed library of native neuro-symbolic
tools shared across all three benchmarks. These tools provide visual
grounding, 3D reconstruction, pose estimation, metric-scale recovery,
text recognition, symbolic camera-layout analysis, and executable
numerical reasoning. A Tool-Use skill organizes calls to these native
tools into a task-specific pipeline.

\begin{table}[H]
    \centering
    \footnotesize
    \setlength{\tabcolsep}{4pt}
    \renewcommand{\arraystretch}{1.12}
    \caption{Native tool atoms available to the agent.}
    \label{tab:native_tools}
    \begin{tabularx}{\columnwidth}{
        @{}
        >{\raggedright\arraybackslash}p{0.46\columnwidth}
        >{\raggedright\arraybackslash}X
        @{}
    }
        \toprule
        \rowcolor{cardtitle}
        \textbf{Tool atom} & \textbf{Function} \\
        \midrule
        \rowcolors{2}{white}{gray!4}

        {\ttfamily\scriptsize
        SemanticDetector.detect}
        &
        Localizes named objects or regions and returns labeled 2D
        bounding boxes.
        \\

        {\ttfamily\scriptsize
        GeometricReconstructor.\par
        reconstruct}
        &
        Recovers camera extrinsics, dense 3D points, confidence maps,
        and depth from one or more images.
        \\

        {\ttfamily\scriptsize
        GeometricReconstructor.\par
        project\_box\_to\_3d\_points}
        &
        Lifts a selected 2D bounding box into its corresponding points
        in the reconstructed 3D scene.
        \\

        {\ttfamily\scriptsize
        ObjPoseEstimator.\par
        predict\_obj\_pose}
        &
        Estimates an object-centric position and orientation from a
        selected object box.
        \\

        {\ttfamily\scriptsize
        MetricScaleEstimator.\par
        estimate\_scale}
        &
        Estimates the conversion from reconstruction coordinates to
        metric units.
        \\

        {\ttfamily\scriptsize
        EasyOCR.ocr}
        &
        Reads visible text that may provide identity, ordering, or
        metric cues.
        \\

        {\ttfamily\scriptsize
        LanguageToCamera.\par
        visualize\_camera\_layout}
        &
        Converts caller-provided numeric view angles and labels into a
        symbolic camera-layout representation.
        \\

        {\ttfamily\scriptsize
        PythonTool.code}
        &
        Generates and executes Python code over workspace variables for
        deterministic geometric computation.
        \\

        \bottomrule
    \end{tabularx}
\end{table}

Tool outputs are stored as typed workspace variables and can be consumed
by later steps. For example, a detector output can be passed to
3D projection, the projected points can be combined with camera
extrinsics, and a geometry skill can use these variables to compute the
final spatial decision.

The visual foundation models supporting these tools are treated as
internal backends rather than planner-visible tool atoms. Specifically,
GroundingDINO supports open-vocabulary detection, VGGT supports
multi-view reconstruction, MoGe supports metric-scale estimation, and
the object-pose backend provides orientation evidence. Fixing these
backends ensures that changing the LVLM backbone does not simultaneously
change the underlying visual tool implementation.

We additionally use \texttt{Planner.reason} as an internal semantic
transition for role binding, option interpretation, and skill retrieval. Since it does not invoke an external perception or computation
module, it is not counted as a native tool execution.

\subsection{Skill Representation and Retrieval}
\label{app:skill_representation}

The skill library contains two complementary types of reusable
experience: \emph{Tool-Use skills} and \emph{geometry skills}.
A Tool-Use skill represents a structured pipeline composed of executable
tool atoms, whereas a geometry skill stores a reusable computational
kernel together with its applicability conditions. Geometry skills are
internally represented as \texttt{python} code lines.

\begin{table}[H]
    \centering
    \footnotesize
    \setlength{\tabcolsep}{3.2pt}
    \renewcommand{\arraystretch}{1.08}
    \caption{Overview of the two reusable skill types.}
    \label{tab:skill_type_overview}
    \begin{tabularx}{\columnwidth}{
        @{}
        >{\raggedright\arraybackslash}p{0.19\columnwidth}
        >{\raggedright\arraybackslash}X
        >{\raggedright\arraybackslash}X
        @{}
    }
        \toprule
        \rowcolor{cardtitle}
        \textbf{Aspect} & \textbf{Tool-Use skill} & \textbf{Geometry skill} \\
        \midrule
        Represents & Executable evidence-collection pipeline
                   & Deterministic numerical decision kernel \\
        Retrieved for & Tool ordering and semantic-role completion
                      & Reference-frame computation and option selection \\
        Stored as & Ordered atoms, bindings, and a stopping condition
                  & Python helpers, usage constraints, and a decision function \\
        Produces & Structured workspace evidence
                 & A serializable decision or diagnostic \\
        \bottomrule
    \end{tabularx}
\end{table}

\subsubsection{Tool-Use Skill Representation}

A Tool-Use skill is stored as a high-level pipeline. Its
\texttt{docstring} describes when the skill should be retrieved, while
\texttt{steps} specify the ordered tool atoms and their semantic
responsibilities. The \texttt{stopping\_condition} defines when the
pipeline has collected sufficient evidence to return control to the
planner.

Listing~\ref{lst:pipeline_skill} presents an abridged pipeline skill for
a single-anchor spatial reasoning problem.

\begin{jsoncard}{Tool-Use Skill Representation}{lst:pipeline_skill}
{
  "id": "mindcube_single_anchor_grounding_skeleton",
  "rule_type": "Tool-Use",
  "level": "high_level",
  "docstring": "Use for viewpoint or movement questions that
    require grounding one named target object.",
  "steps": [
    {
      "atom": "GeometricReconstructor.reconstruct",
      "instruction": "Recover the shared scene geometry and
        camera poses."
    },
    {
      "atom": "SemanticDetector.detect",
      "instruction": "Detect the named target object."
    },
    {
      "atom": "GeometricReconstructor.project_box_to_3d_points",
      "instruction": "Project the selected target box into the
        shared 3D frame."
    },
    {
      "atom": "PythonTool.code",
      "instruction": "Compute the requested spatial relation in
        the specified camera frame."
    }
  ],
  "stopping_condition": "The target has been grounded in 3D and
    the requested relation can be deterministically evaluated.",
  "status": "seed"
}
\end{jsoncard}

The pipeline representation separates reusable structure from
instance-specific bindings. For example, the skill specifies that a
target must be detected and projected, while the target name, source
view, selected bounding box, reference camera, and answer-option mapping
are bound from the current question at execution time.

\subsubsection{Geometry Skill Representation}

A geometry skill is represented as a self-contained computational
kernel. Internally, the kernel may contain multiple helper routines,
each implementing an atomic geometric operation, such as robust point
aggregation, coordinate transformation, option parsing, or directional
scoring. These routines are composed into a high-level decision
function and stored as a single Geometry skill.

Listing~\ref{lst:geometry_skill} shows the internal composition of a
representative geometry skill. For readability, metadata are rendered as
comments and the executable kernel is shown as ordinary Python; the actual
record serializes the same lines in its \texttt{code\_lines} field. Function
bodies are abbreviated, while the decomposition follows the stored rule.

\begin{pythoncard}{Geometry Skill Kernel}{lst:geometry_skill}
# id: camera_frame_object_quadrant_choice
# rule_type: Geometry; level: high_level
# usage:
# - Require camera extrinsics, a reference-camera index,
#   grounded target points, and an explicit option mapping.
# - Interpret +X as right and +Z as forward.
# - Return a diagnostic when geometry or option semantics
#   are missing or ambiguous.

def _to_numpy(value):
    # Convert tensors or array-like inputs to NumPy.
    ...

def _normalize(vector):
    # Validate and normalize a direction vector.
    ...

def _extract_points_and_weights(target_points):
    # Unpack 3D points and optional confidence weights.
    ...

def _robust_center(points, weights=None):
    # Estimate a confidence-weighted or median center.
    ...

def _world_to_camera(point, extrinsic):
    # Transform a world point into camera coordinates.
    ...

def _parse_option_direction(option):
    # Convert option semantics into an XZ direction.
    ...

def reusable_decision(reconstruction, ref_camera_index,
                      option_vectors, target_points_world):
    points, weights = _extract_points_and_weights(
        target_points_world)
    center_world = _robust_center(points, weights)
    center_camera = _world_to_camera(
        center_world,
        reconstruction.extrinsic[ref_camera_index])
    target_direction = _normalize(
        [center_camera[0], center_camera[2]])
    option_directions = {
        key: _parse_option_direction(value)
        for key, value in option_vectors.items()
    }
    scores = score_options(
        target_direction, option_directions)
    return select_best_option_or_diagnostic(scores)

# status: active
# empirical_stats: seen=145, matched=39, used=39,
#                  correct_after_use=24,
#                  incorrect_after_use=15
\end{pythoncard}

The helper routines provide a modular decomposition of the geometry
kernel:

\begin{itemize}
    \item \texttt{\_extract\_points\_and\_weights} binds projection
    outputs to the expected numerical inputs;
    \item \texttt{\_robust\_center} aggregates noisy 3D observations;
    \item \texttt{\_world\_to\_camera} enforces the requested reference
    frame;
    \item \texttt{\_parse\_option\_direction} converts answer semantics
    into comparable geometric vectors; and
    \item \texttt{reusable\_decision} composes these operations into an
    auditable high-level decision kernel.
\end{itemize}

These helper routines are not independently retrieved by the skill
selector. Instead, the selector retrieves the high-level geometry skill
through its docstring, after which the complete kernel and its usage
constraints are exposed to the reasoning model.

The \texttt{empirical\_stats} field is updated from subsequent
trajectories. Here, \texttt{matched} denotes retrieval by the selector,
whereas \texttt{used} indicates that the skill was actually executed.
The outcome counters record whether samples answered after using the
skill were judged correct or incorrect. These statistics support
empirical promotion and retirement, but are not exposed to the
first-stage selector.

\subsubsection{Two-Stage Skill Retrieval}

Providing every complete skill to the reasoning model would introduce
substantial context length and irrelevant executable details.
We therefore use two-stage retrieval. In the first stage, each pipeline
is reduced to a compact selection payload containing its identifier,
type, level, docstring, and a bounded structural hint.

\begin{jsoncard}{Compact Selector Payload}{lst:selector_payload}
{
  "id": "mindcube_single_anchor_grounding_skeleton",
  "rule_type": "pipeline",
  "level": "high_level",
  "docstring": "Use for viewpoint or movement questions that
    require grounding one named target object.",
  "structure_hint": "GeometricReconstructor.reconstruct >
    SemanticDetector.detect >
    GeometricReconstructor.project_box_to_3d_points >
    PythonTool.code | roles=target,anchor,camera"
}
\end{jsoncard}

Given the question, a compact workspace summary, and these selection
payloads, the selector returns only the identifiers of relevant skills:

\begin{jsoncard}{Skill-Selector Output}{lst:selector_output}
{
  "selected_rule_ids": [
    "mindcube_single_anchor_grounding_skeleton"
  ],
  "rationale": "The question names one target and requires
    evaluating its relation in a specified viewpoint."
}
\end{jsoncard}

In the second stage, the complete stored representation is expanded
only for the selected skills. For a Tool-Use skill, the planner receives
the ordered atoms, step instructions, and stopping condition. For a
geometry skill, it receives the usage constraints and reusable code
kernel. This design preserves semantic and executable information after
selection while keeping the initial retrieval context compact.

%% file: Appendix/B.tex
\section{Case Studies}
\label{app:case_studies}

\subsection{End-to-End Tool Execution}
\label{app:end_to_end_cases}

We present representative examples to illustrate how the agent retrieves
skills, executes native tools, reuses geometry kernels, and derives its
final answer from explicit tool outputs.

\subsubsection{Case 1: Camera-Motion Reasoning on MindCube}
\label{app:case_mindcube_motion}

This case illustrates how a Tool-Use skill and a Geometry skill work
together to solve a camera-motion question. The input consists of two
views of the same scene, and the agent must determine the motion of the
camera from the first view to the second.

\casestep{Input.}
The question asks:

\begin{figure*}[t]
    \centering
    \begin{minipage}{0.47\textwidth}
        \centering
        \includegraphics[
            width=\linewidth
        ]{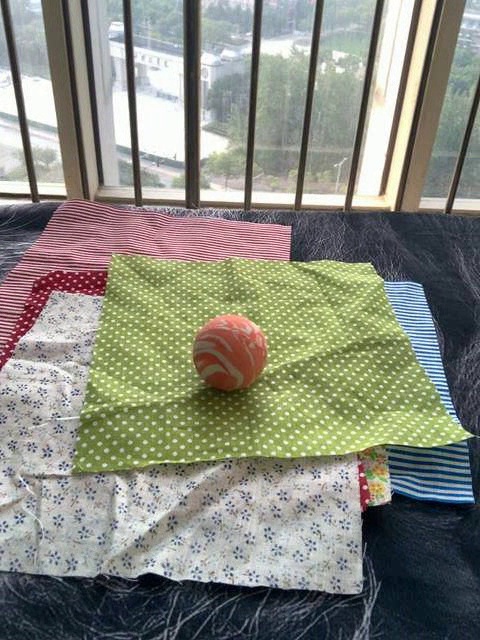}

        \small (a) First input view
    \end{minipage}
    \hfill
    \begin{minipage}{0.47\textwidth}
        \centering
        \includegraphics[
            width=\linewidth
        ]{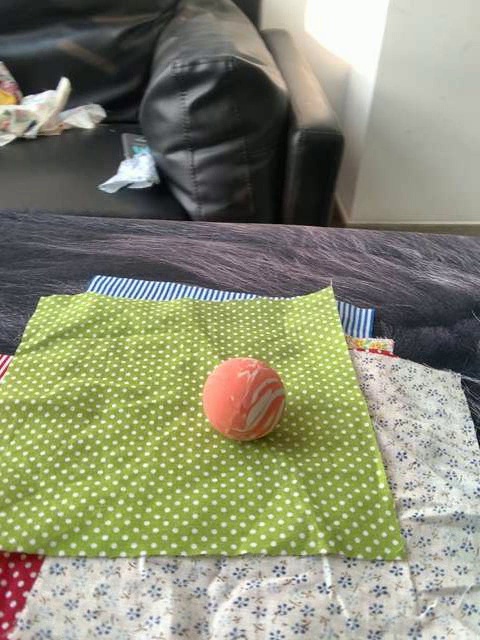}

        \small (b) Second input view
    \end{minipage}
    \caption{
        Original input images for the MindCube camera-motion
        case. The task asks for the camera motion from the first
        view to the second view.
    }
    \label{fig:case1_input}
\end{figure*}

\begin{questionbox}
\textbf{Question.}
Based on these two views in Figure~\ref{fig:case1_input} showing the same
scene, in which direction did I move from the first view to the second
view?

\smallskip
\textbf{Options.}
(A) diagonally forward and right; (B) directly right;
(C) diagonally forward and left; and (D) directly left.
\end{questionbox}

\casestep{Skill Retrieval.}
The selector first examines the compact descriptions of the available
Tool-Use skills. Since the question requires reasoning about the
relative motion between two cameras, it retrieves
\path{mindcube_recon_then_python_skeleton}. After selection, the
complete pipeline is exposed to the agent.

\begin{skillcard}{Retrieved Tool-Use Skill}

\skillid{mindcube_recon_then_python_skeleton}
\textbf{Applicability.}
Use when a multi-view reconstruction directly contains the camera
geometry required by the question and no explicit object grounding is
needed.

\medskip
\textbf{Step 1: Reconstruct the shared scene.}

Call \path{GeometricReconstructor.reconstruct} on all input images to
recover their shared geometry and camera poses.

\begin{center}
$\boldsymbol{\downarrow}$
\end{center}

\textbf{Step 2: Resolve the spatial relation.}

Pass the reconstruction output to \path{PythonTool.code} and use a
Geometry skill to compute the camera-frame motion and map it to one of
the answer choices.

\medskip
\textbf{Stopping condition.}
Stop when the recovered geometry supports a unique
benchmark-formatted answer without additional object grounding.

\end{skillcard}

The retrieved skill determines the high-level execution order, while
leaving the benchmark-specific geometric computation to the subsequent
Geometry skill.

\casestep{Scene Reconstruction.}
Following the retrieved pipeline, the agent invokes
\path{GeometricReconstructor.reconstruct} on the two input images. The
tool successfully produces a shared reconstruction and estimates the
extrinsic parameters of both cameras.

This operation converts the original visual question into a structured
geometric problem. The camera motion can now be computed in a common
coordinate system instead of being inferred directly from appearance.

\casestep{Geometry Reasoning.}
Once the camera poses are available, the selector retrieves
\path{camera0_motion_option_from_extrinsics}. The skill defines the
first camera as the reference frame and interprets the displacement of
the second camera relative to it.

The following listing shows the core structure of the retrieved
Geometry skill. Low-level validation and diagnostic statements are
omitted for clarity.

\begin{codecard}{CAMERA-MOTION GEOMETRY KERNEL}
# Geometry skill:
# camera0_motion_option_from_extrinsics
def reusable_decision(extrinsic, options_text):
    ext = normalize_extrinsics(extrinsic)

    center_0 = camera_center(ext[0])
    center_1 = camera_center(ext[1])

    motion_world = center_1 - center_0
    motion_cam0 = to_camera0_frame(
        motion_world, ext[0]
    )

    direction = normalize_xz(motion_cam0)
    options = parse_motion_options(options_text)

    return select_best_direction(
        direction, options
    )
\end{codecard}

The helper routines implement atomic geometric operations, including
extrinsic normalization, camera-center recovery, reference-frame
conversion, option parsing, and direction comparison. The high-level
Geometry skill composes these operations into a reusable decision
kernel.

For this instance, the kernel identifies both a leftward component and
a forward component in the second camera's motion.

\casestep{Final Decision.}
\begin{resultbox}
\textbf{Result.} The recovered motion is diagonally forward-left,
corresponding to option~\textbf{C}, which agrees with the ground-truth
answer.
\end{resultbox}

\medskip
\noindent
\emph{Execution path:}
two-view input
$\rightarrow$ Tool-Use skill retrieval
$\rightarrow$ scene reconstruction
$\rightarrow$ Geometry skill retrieval
$\rightarrow$ numerical execution
$\rightarrow$ \textbf{C}.

\subsubsection{Case 2: Grounded Object-Direction Reasoning on MMSI}
\label{app:case_mmsi_grounding}

This case requires both visual grounding and camera-frame geometric
reasoning. Unlike Case 1, the target
location cannot be obtained from the camera poses alone. The agent must
first identify the specified object and lift its image-space location
into the reconstructed 3D scene.

\casestep{Input.}
The question asks:
\begin{figure*}[t]
    \centering
    \begin{minipage}{0.47\textwidth}
        \centering
        \includegraphics[
            width=\linewidth
        ]{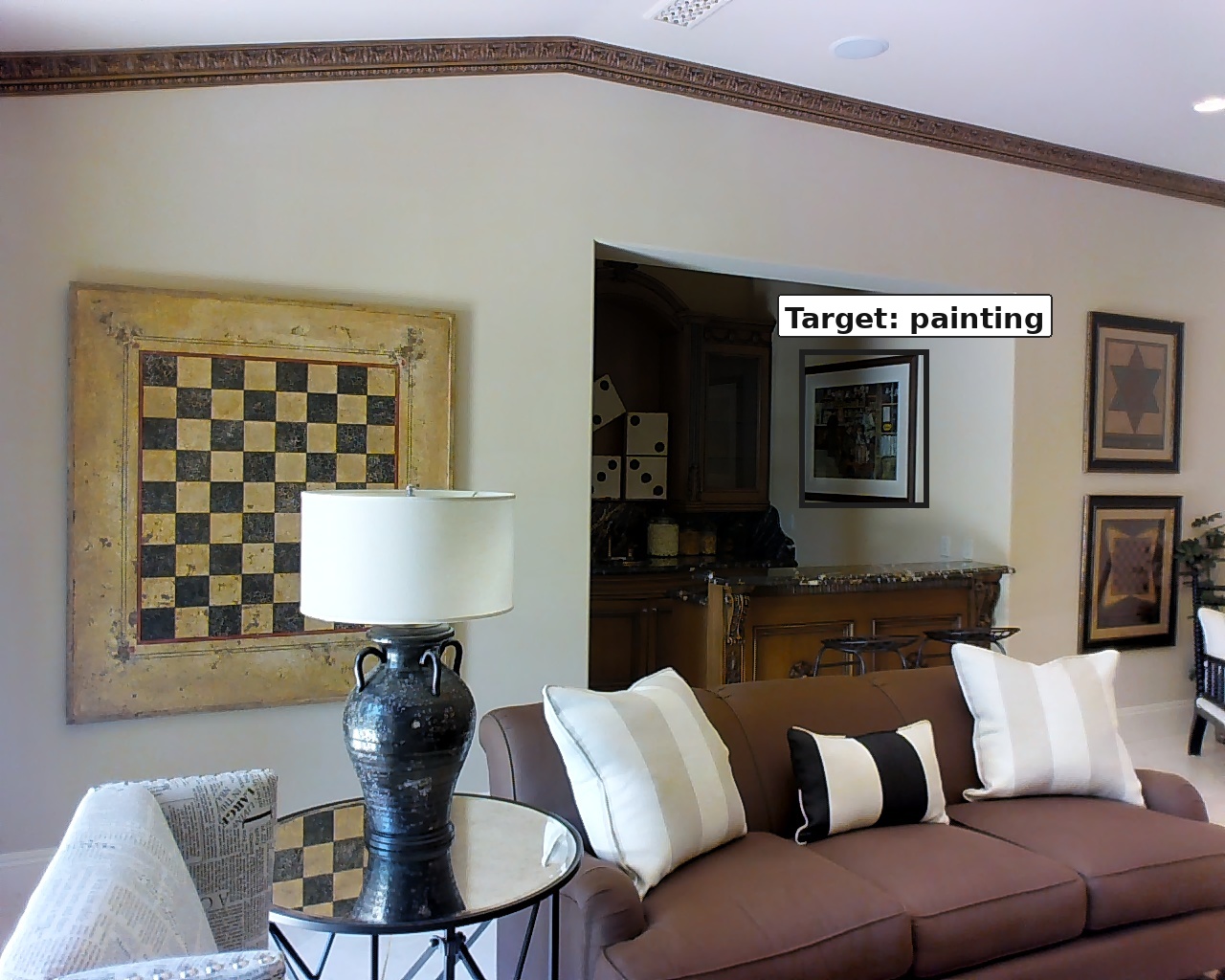}

        \small (a) First input view
    \end{minipage}
    \hfill
    \begin{minipage}{0.47\textwidth}
        \centering
        \includegraphics[
            width=\linewidth
        ]{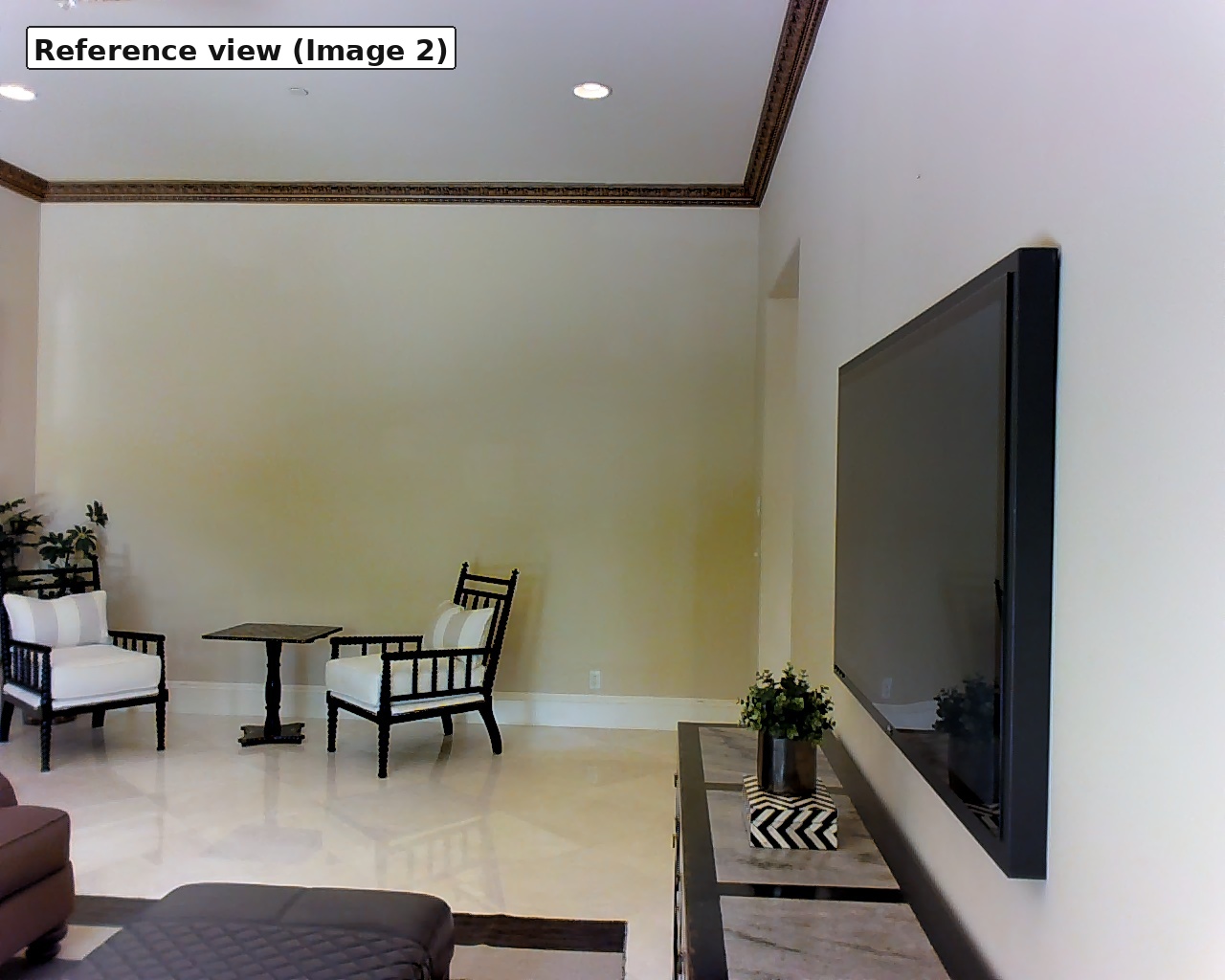}

        \small (b) Second input view
    \end{minipage}
    \caption{
        Original input images for the MMSI grounded-direction
        case. The painting is grounded in Image 1, while its
        direction is evaluated relative to the observer in Image 2.
    }
    \label{fig:case2_input}
\end{figure*}

\begin{questionbox}
\textbf{Question.}
When you took the photo in Image 2 in Figure~\ref{fig:case2_input}, where
was the painting with a black frame and white background in relation to
you?

\smallskip
\textbf{Options.}
(A) rear right; (B) rear left; (C) front left; and
(D) directly to the left.
\end{questionbox}

\casestep{Skill Retrieval.}
The question names one target object and asks for its position relative
to the observer in a specified image. The selector therefore retrieves
\path{mmsi_single_anchor_grounding_skeleton}.

\begin{skillcard}{Retrieved Tool-Use Skill}
\skillid{mmsi_single_anchor_grounding_skeleton}
\textbf{Applicability.}
Use when a question asks where one named object or scene region is
located relative to an observer or a camera-defined reference frame.

\medskip
\textbf{Step 1: Reconstruct the scene.}

Call \path{GeometricReconstructor.reconstruct} to recover a shared 3D
scene and the camera poses of the input images.

\begin{center}
$\boldsymbol{\downarrow}$
\end{center}

\textbf{Step 2: Ground the target.}

Call \path{SemanticDetector.detect} to localize the named object in an
input image.

\begin{center}
$\boldsymbol{\downarrow}$
\end{center}

\textbf{Step 3: Lift the target into 3D.}

Call
\path{GeometricReconstructor.project_box_to_3d_points}
to project the selected detection into the shared 3D scene.

\begin{center}
$\boldsymbol{\downarrow}$
\end{center}

\textbf{Step 4: Resolve the camera-frame relation.}

Pass the reconstruction and grounded target points to
\path{PythonTool.code}. Apply a Geometry skill to express the target
relative to the requested camera and select the corresponding option.

\medskip
\textbf{Stopping condition.}
Stop when the target has been grounded in 3D and its direction in the
requested camera frame maps to one unique answer.

\end{skillcard}

The retrieved skill determines both the required tool sequence and the
data flow between the tools: the detector supplies a bounding box to
the projection tool, and the projected points are subsequently consumed
by the geometric decision kernel.

\casestep{Scene Reconstruction.}
Following the retrieved pipeline, the agent first invokes
\path{GeometricReconstructor.reconstruct} on the two images. The
operation successfully recovers a common 3D scene and estimates the
poses of both cameras.

The resulting reconstruction provides the transformation required to
express an object observed in Image 1 relative to the observer in
Image 2.

\casestep{Target Grounding.}
The agent invokes \path{SemanticDetector.detect} with the description
``the painting with a black frame and white background.'' The detector
returns multiple candidate boxes. By inspecting their visual locations,
the agent selects the small framed artwork near the bar opening in
Image 1.

The selected box is then passed to
\path{GeometricReconstructor.project_box_to_3d_points}. This operation
lifts the painting from its two-dimensional bounding box into a set of
points in the shared 3D coordinate frame.

At this stage, the workspace contains both pieces of evidence required
for geometric reasoning: the Image 2 camera pose and the 3D points
associated with the painting.

\casestep{Geometry Reasoning.}
The selector retrieves
\path{camera_frame_object_quadrant_choice}. This skill determines the
direction of a grounded object relative to a selected camera and maps
that direction to the semantic labels in the answer choices.

The core structure of the retrieved Geometry skill is shown below.
Input validation and diagnostic branches are omitted for clarity.

\begin{codecard}{CAMERA-FRAME OBJECT-QUADRANT KERNEL}
# Geometry skill:
# camera_frame_object_quadrant_choice

def reusable_decision(
    reconstruction,
    ref_camera_index,
    option_vectors,
    target_points_world,
    target_point_weights
):
    center_world = robust_weighted_center(
        target_points_world,
        target_point_weights
    )

    center_camera = world_to_camera(
        center_world,
        reconstruction.extrinsic[
            ref_camera_index
        ]
    )

    direction = normalize_xz(center_camera)
    options = parse_direction_options(
        option_vectors
    )

    return select_best_direction(
        direction, options
    )
\end{codecard}

For this instance, the agent binds the reference camera to Image 2 and
passes the projected painting points as the target. The helper
operations estimate a robust target center, transform it from the
shared world frame into the Image 2 camera frame, and compare its
planar direction with the four candidate relations.

The resulting camera-frame direction is front-left.

\casestep{Final Decision.}
\begin{resultbox}
\textbf{Result.} The grounded painting is front-left in the Image~2
camera frame, corresponding to option~\textbf{C}.
\end{resultbox}

\medskip
\noindent
\emph{Execution path:}
two-image input
$\rightarrow$ Tool-Use skill retrieval
$\rightarrow$ scene reconstruction
$\rightarrow$ target detection
$\rightarrow$ 3D projection
$\rightarrow$ Geometry skill retrieval
$\rightarrow$ camera-frame reasoning
$\rightarrow$ \textbf{C}.

\subsubsection{Case 3: Allocentric Relation Reasoning on OmniSpatial}
\label{app:case_omnispatial_allocentric}

This case demonstrates allocentric spatial reasoning, in which the
required reference frame belongs to a visible entity rather than to
the camera. The agent must estimate the orientation of a projection
screen and determine the position of an electric fan from the screen's
viewpoint.

\casestep{Input.}
The question asks:

\begin{figure}[t]
    \centering
    \includegraphics[
        width=0.92\columnwidth,
        trim=0 85 0 165,
        clip
    ]{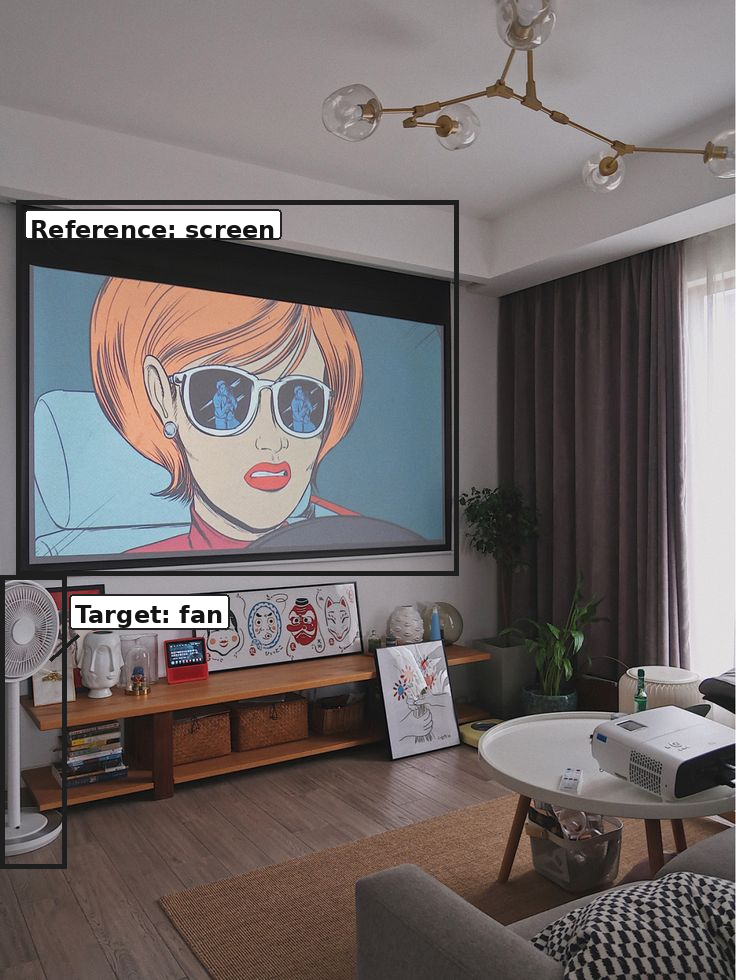}
    \caption{
        Original input image for the OmniSpatial allocentric
        case. The electric fan is localized relative to the
        projection screen's object-centric viewpoint.
    }
    \label{fig:case3_input}
\end{figure}

\begin{questionbox}
\textbf{Question.}
In Figure~\ref{fig:case3_input}, where is the electric fan, seen from the
projection screen's viewpoint?

\smallskip
\textbf{Options.}
(A) right and (B) left.
\end{questionbox}

\casestep{Skill Retrieval.}
The selector identifies the projection screen as the reference entity,
the electric fan as the target entity, and left--right as the requested
relation. It retrieves
\path{omnispatial_allocentric_pairwise_role_completion_skeleton}.

\begin{skillcard}{Retrieved Tool-Use Skill}
\skillid{omnispatial_allocentric_pairwise_role_completion_skeleton}
\textbf{Applicability.}
Use for a visible-entity viewpoint question containing one oriented
reference entity and one target entity, where the target must be
classified in the reference entity's coordinate frame.

\medskip
\textbf{Step 1: Bind the semantic roles.}

Identify the reference entity, target entity, requested relation, and
option meanings separately.

\begin{center}
$\boldsymbol{\downarrow}$
\end{center}

\textbf{Step 2: Reconstruct the scene.}

Call \path{GeometricReconstructor.reconstruct} to establish a shared
3D coordinate frame.

\begin{center}
$\boldsymbol{\downarrow}$
\end{center}

\textbf{Step 3: Ground both entities.}

Call \path{SemanticDetector.detect} separately for the reference and
target. One aggregate detection cannot fill both roles.

\begin{center}
$\boldsymbol{\downarrow}$
\end{center}

\textbf{Step 4: Estimate the reference orientation.}

Call \path{ObjPoseEstimator.predict_obj_pose} on the selected reference
box to establish its object-centric frame.

\begin{center}
$\boldsymbol{\downarrow}$
\end{center}

\textbf{Step 5: Lift both entities into 3D.}

Call
\path{GeometricReconstructor.project_box_to_3d_points}
for both selected boxes in the same reconstruction.

\begin{center}
$\boldsymbol{\downarrow}$
\end{center}

\textbf{Step 6: Resolve the allocentric relation.}

Pass the reference pose and both 3D anchors to
\path{PythonTool.code}, then apply a Geometry skill to classify the
target in the reference frame.

\medskip
\textbf{Stopping condition.}
Stop when the reference identity and orientation, target identity,
same-frame 3D anchors, and option mapping jointly support one answer.

\end{skillcard}

This skill is more constrained than a simple visual left--right
heuristic. In particular, it requires explicit orientation evidence
for the reference entity and does not permit the camera frame to be
silently substituted for the reference frame.

\casestep{Scene Reconstruction.}
The agent invokes \path{GeometricReconstructor.reconstruct} on the
input image. The tool recovers a 3D representation of the scene that
will be shared by the subsequent grounding and pose-estimation steps.

\casestep{Role-Specific Grounding.}
The agent invokes \path{SemanticDetector.detect} twice with
role-specific prompts: once for the projection screen and once for the
electric fan. Both calls return one valid detection.

Keeping these calls separate preserves the semantic distinction between
the reference and target. The screen detection defines the entity whose
viewpoint is requested, while the fan detection defines the entity
whose relative position must be determined.

\casestep{Reference-Frame Construction.}
The selected screen box is passed to
\path{ObjPoseEstimator.predict_obj_pose}. The resulting semantic 6D
pose provides the orientation of the screen and defines its local
left--right frame.

The agent then projects both the screen box and the fan box through
\path{GeometricReconstructor.project_box_to_3d_points}. This produces
two point sets in the same reconstructed world frame: one for the
reference screen and one for the target fan.

\casestep{Geometry Reasoning.}
With the reference pose and both 3D anchors available, the selector
retrieves
\path{allocentric_left_right_from_reference_pose}.

The following listing shows the central computation of the retrieved
Geometry skill. Validation and diagnostic branches are omitted for
clarity.

\begin{codecard}{ALLOCENTRIC LEFT--RIGHT GEOMETRY KERNEL}
# Geometry skill:
# allocentric_left_right_from_reference_pose

def decide_allocentric_left_right(
    reference_pose,
    reference_points_3d,
    target_points_3d,
    option_to_side
):
    reference_center = robust_center(
        reference_points_3d
    )
    target_center = robust_center(
        target_points_3d
    )

    object_to_world = (
        reference_pose.T_obj2world.copy()
    )
    object_to_world[:3, 3] = reference_center
    world_to_reference = invert(
        object_to_world
    )

    target_local = transform_point(
        target_center,
        world_to_reference
    )

    side = (
        "right"
        if target_local[0] > 0
        else "left"
    )

    return option_for_side(
        side, option_to_side
    )
\end{codecard}

The skill first estimates robust centers for the screen and fan. It
then combines the screen center with its estimated orientation to
construct a screen-centric coordinate frame. The fan center is
transformed into this frame, and the sign of its local lateral
component determines whether it lies to the screen's left or right.

For this instance, the transformed fan position has a positive
rightward component in the screen-centric frame.

\casestep{Final Decision.}
\begin{resultbox}
\textbf{Result.} The fan lies to the screen's right in the
screen-centric frame, corresponding to option~\textbf{A}.
\end{resultbox}

\medskip
\noindent
\emph{Execution path:}
single-image input
$\rightarrow$ Tool-Use skill retrieval
$\rightarrow$ scene reconstruction
$\rightarrow$ reference and target detection
$\rightarrow$ reference pose estimation
$\rightarrow$ paired 3D projection
$\rightarrow$ Geometry skill retrieval
$\rightarrow$ allocentric transformation
$\rightarrow$ \textbf{A}.

\subsection{Online Skill Evolution}
\label{app:online_skill_evolution}

The preceding cases demonstrate how existing skills guide inference.
We next show how the skill library itself changes during online
interaction. We separately examine the evolution of a Tool-Use skill
and a Geometry skill.

\subsubsection{Case 4: Evolution of a Tool-Use Skill}
\label{app:case_tool_use_evolution}

\begin{figure*}[!t]
    \centering
    \setlength{\tabcolsep}{2pt}

    \begin{tabular}{@{}cccc@{}}
        \includegraphics[
            width=0.235\textwidth,
            height=0.18\textheight,
            keepaspectratio
        ]{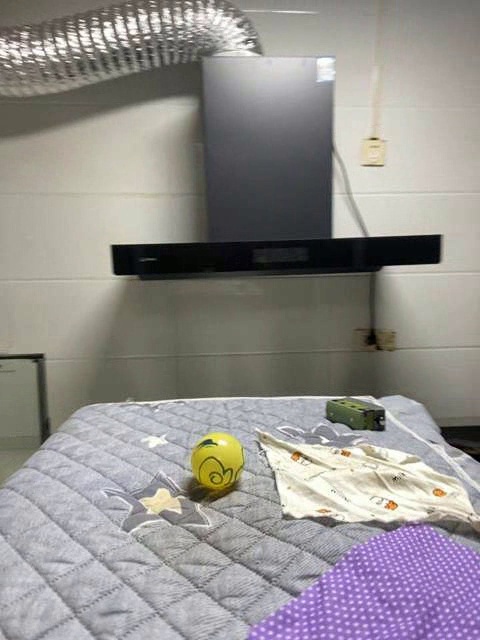}
        &
        \includegraphics[
            width=0.235\textwidth,
            height=0.18\textheight,
            keepaspectratio
        ]{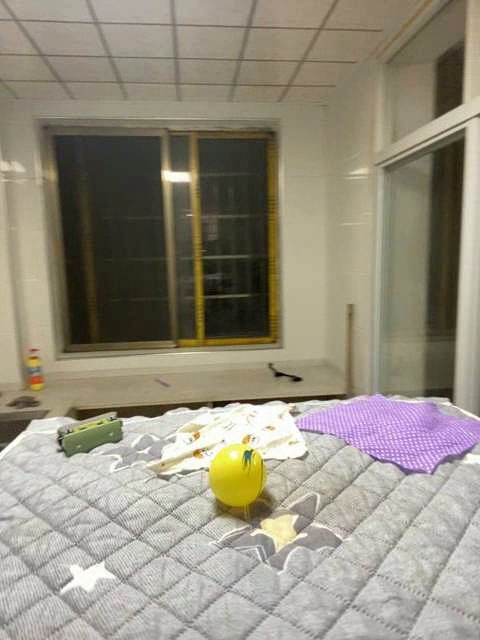}
        &
        \includegraphics[
            width=0.235\textwidth,
            height=0.18\textheight,
            keepaspectratio
        ]{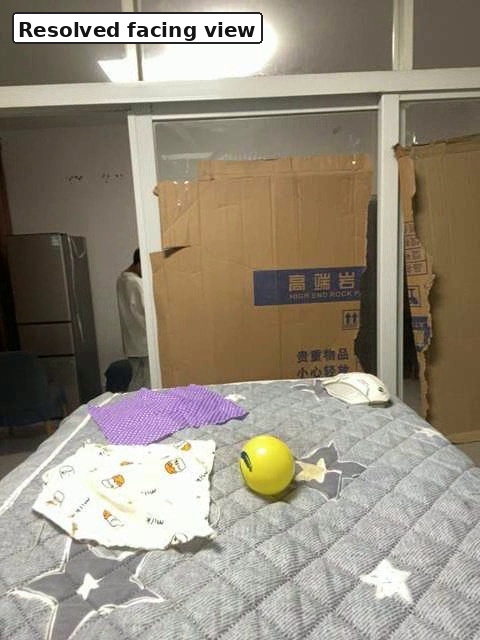}
        &
        \includegraphics[
            width=0.235\textwidth,
            height=0.18\textheight,
            keepaspectratio
        ]{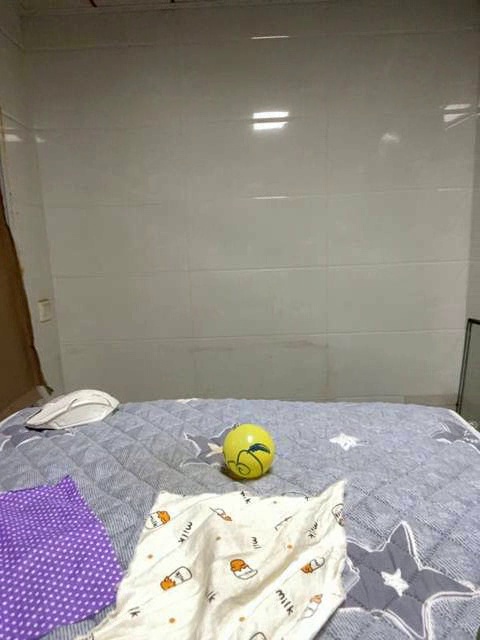}
        \\[-2pt]
        {\scriptsize (a) Front}
        &
        {\scriptsize (b) Left}
        &
        {\scriptsize (c) Back}
        &
        {\scriptsize (d) Right}
    \end{tabular}

    \vspace{-2mm}
    \caption{
        Original four-view input for the evolved Tool-Use skill.
    }
    \label{fig:case4_input}
\end{figure*}

In Figure~\ref{fig:case4_input}, this case concerns four-view MindCube questions in which an observer
starts from a specified viewpoint, turns left or right, moves forward,
and must decide whether the movement approaches a named destination
object.

\casestep{Initial Skill.}
For this task family, the initial library contains
\path{four_view_turn_move_symbolic_layout_root}. The skill reconstructs
the shared scene and resolves the observer's orientation after the
specified turn.

\begin{skillcard}{Initial Tool-Use Skill}

\skillid{four_view_turn_move_symbolic_layout_root}
\textbf{Applicability.}
Use for same-position multi-view questions that provide an ordered
camera layout and ask about turning left or right from a named view.

\medskip
\textbf{Pipeline.}

\begin{center}
\path{GeometricReconstructor.reconstruct}

\(\boldsymbol{\downarrow}\)

\path{LanguageToCamera.visualize_camera_layout}
\end{center}

\textbf{Stopping condition.}
Stop when the shared scene and the observer's post-turn orientation
have been recovered.

\end{skillcard}

This pipeline is sufficient for questions that only ask which direction
the observer faces after turning. However, it does not fully solve
questions that additionally ask whether moving forward approaches a
specific destination object.

\casestep{Observed Failure.}
In unsuccessful trajectories, the initial pipeline correctly
reconstructed the scene and resolved the post-turn viewpoint, but then
stopped. Consequently, the workspace contained the observer's final
orientation but not the location of the named destination.

For example, after resolving a left turn from Image 2, the agent could
determine that the observer would face the direction represented by
Image 3. However, it had not yet grounded the target object in Image 3
and therefore could not determine whether forward movement would reduce
the distance to that object.

The missing evidence was thus not another camera-layout operation. It
was a destination-grounding and movement-comparison stage after the
symbolic turn had been resolved.

\casestep{Residual Induction.}
The online induction process groups successful and unsuccessful
trajectories from the same task family. Both groups share the following
parent prefix:

\begin{center}
\small
scene reconstruction
\(\rightarrow\)
symbolic turn resolution.
\end{center}

Successful trajectories continue with two additional semantic
operations:

\begin{center}
\small
ground the named destination in the turned-facing view
\(\rightarrow\)
evaluate whether forward movement approaches it.
\end{center}

In contrast, the unsuccessful trajectories terminate after the shared
prefix. This comparison identifies the latter two operations as a
reusable residual extension of the parent pipeline.

Importantly, the induction process uses more than the atom sequence.
The trace records specify that the detector must ground the
\emph{destination entity} in the \emph{turned-facing image}, and that
the numerical step must evaluate distance change in the
\emph{post-turn observer frame}.

\casestep{Branch Fusion.}
Fusion preserves the original pipeline and creates a specialized branch
for turn-then-move questions that mention a destination object. The
new branch is stored as
\path{four_view_turn_move_target_grounding_branch}.

\begin{skillcard}{Tool-Use Skill After Online Fusion}

\skillid{four_view_turn_move_target_grounding_branch}
\textbf{Parent skill.}

\path{four_view_turn_move_symbolic_layout_root}

\medskip
\textbf{Retained parent prefix.}

\begin{center}
\path{GeometricReconstructor.reconstruct}

\(\boldsymbol{\downarrow}\)

\path{LanguageToCamera.visualize_camera_layout}
\end{center}

\hrule
\medskip

\textbf{Online extension.}

\begin{center}
\path{SemanticDetector.detect}

\(\boldsymbol{\downarrow}\)

\path{PythonTool.code}
\end{center}

The detector grounds the named destination in the image corresponding
to the resolved post-turn orientation. The numerical step then
determines whether forward motion in that frame decreases the distance
to the grounded destination.

\medskip
\textbf{New stopping condition.}
Stop only after the turned-facing view, destination grounding, and
forward-motion comparison jointly produce a yes--no answer.

\end{skillcard}

The fusion operation therefore does not overwrite the more general
parent skill. The original pipeline remains applicable to pure
viewpoint questions, while the new branch handles the narrower
turn-then-move task family.

\casestep{Later Reuse.}
In a subsequent query, the observer starts from Image 2, turns left,
moves forward, and asks whether the movement gets closer to a smoking
machine. The selector retrieves the evolved branch and instantiates
its open semantic slots using the current question.

\begin{skillcard}{Instantiated Evolved Tool-Use Skill}
\skillid{four_view_turn_move_target_grounding_branch}
\textbf{Current semantic bindings.}

\smallskip
Starting viewpoint: Image 2

Turn operation: left turn

Resolved facing view: Image 3

Destination entity: smoking machine

Requested decision: whether forward movement approaches the
destination

\medskip
\textbf{Instantiated execution.}

\smallskip
\textbf{1. Shared-scene reconstruction}

\path{GeometricReconstructor.reconstruct}

Recover the shared geometry and camera poses from the four ordered
views.

\begin{center}
$\boldsymbol{\downarrow}$
\end{center}

\textbf{2. Post-turn viewpoint resolution}

\path{LanguageToCamera.visualize_camera_layout}

Bind Image 2 as the starting viewpoint and apply a left turn. The
resulting forward direction is aligned with Image 3.

\begin{center}
$\boldsymbol{\downarrow}$
\end{center}

\textbf{3. Destination grounding}

\path{SemanticDetector.detect}

Detect the smoking machine specifically in Image 3, which is the
resolved post-turn reference view.

\begin{center}
$\boldsymbol{\downarrow}$
\end{center}

\textbf{4. Forward-motion comparison}

\path{PythonTool.code}

Use the reconstruction, resolved camera layout, and grounded
destination to determine whether moving forward decreases the distance
to the smoking machine.

\medskip
\textbf{Output.}

The forward movement approaches the destination:
\textbf{A. Yes}.

\end{skillcard}

The first generated numerical program contains an execution-level
syntax error. The agent regenerates the computation while preserving
the retrieved pipeline, semantic bindings, and reference frame. The
corrected execution returns \textbf{A. Yes}.

\medskip
\noindent
\emph{Evolution path:}
partial parent pipeline
$\rightarrow$ recurring missing destination evidence
$\rightarrow$ trace-level residual induction
$\rightarrow$ candidate branch fusion
$\rightarrow$ instantiated evolved skill
$\rightarrow$ successful execution.

\subsubsection{Case 5: Evolution of a Geometry Skill}
\label{app:case_geometry_evolution}

This case shows how repeated sample-specific computations are converted
into a reusable Geometry skill. In figure~\ref{fig:case1_input}, the task family consists of two-view
camera-motion questions that ask whether the second camera moved left,
right, diagonally forward-left, or diagonally forward-right relative
to the first camera.

\casestep{Initial Execution.}
The initial Tool-Use skill already provides the appropriate high-level
pipeline:

\begin{center}
\small
scene reconstruction
\(\rightarrow\)
numerical reasoning over camera extrinsics.
\end{center}

However, the initial library does not contain a specialized Geometry
skill for converting two world-to-camera extrinsics into one of the
benchmark motion choices. Consequently, \path{PythonTool.code} must
generate this computation independently for each sample.

Different trajectories implement the same underlying operations in
slightly different ways: invert camera extrinsics, recover camera
centers, compute the displacement between the views, and compare its
lateral and forward components with the answer choices.

\casestep{Recurring Computation.}
The online trace buffer contains multiple two-view motion trajectories
with the same semantic objective and data requirements.

A successful trajectory correctly recovers the second camera center,
expresses its displacement relative to the first camera, and selects
the corresponding diagonal motion option. Another trajectory attempts
the same computation but uses brittle option parsing and fallback
behavior. When the current options are not extracted correctly, its
sample-specific program can return an unrelated or incorrectly mapped
label.

The two trajectories therefore reveal both the reusable computation
and the behavior that must not be retained:

\begin{skillcard}{Recurring Geometry Pattern}
\textbf{Required inputs.}

A stack containing at least two world-to-camera extrinsic matrices and
a set of answer choices describing left, right, or diagonal-forward
camera motion.

\medskip
\textbf{Shared computation.}

\begin{center}
recover camera centers

\(\boldsymbol{\downarrow}\)

express camera-1 displacement in the camera-0 frame

\(\boldsymbol{\downarrow}\)

retain the lateral and forward components

\(\boldsymbol{\downarrow}\)

parse the current option semantics

\(\boldsymbol{\downarrow}\)

select the best-supported motion option
\end{center}

\textbf{Observed failure mode.}

A sample-specific implementation may silently use hard-coded option
labels or a default answer when option parsing, extrinsic validation,
or motion estimation fails.

\end{skillcard}

Because the traces share the same reference frame, input schema,
geometric transformation, and output semantics, they form one
code-control family rather than unrelated Python programs.

\casestep{Atomic-Operation Induction.}
The induction process decomposes the recurring programs into a small
set of reusable geometric operations:

\begin{itemize}
    \item normalize and validate the extrinsic stack;
    \item recover a camera center from a world-to-camera transform;
    \item express camera displacement in the first-camera frame;
    \item parse direction semantics from the current answer choices;
    \item score direct and diagonal motion candidates; and
    \item return a diagnostic instead of fabricating an answer when
          the required evidence is invalid.
\end{itemize}

These operations are not stored as independently retrievable skills.
They are composed as helper routines inside one self-contained
high-level Geometry kernel.

\casestep{Geometry-Skill Fusion.}
Fusion produces the candidate Geometry skill
\path{camera0_motion_option_from_extrinsics}. The skill records not
only the reusable code but also its applicability conditions, input
bindings, reference-frame convention, and failure guards.

\begin{skillcard}{Evolved Geometry Skill}
\skillid{camera0_motion_option_from_extrinsics}
\textbf{Function.}

Choose among direct-left, direct-right, diagonal-forward-left, and
diagonal-forward-right options by expressing the second camera's
motion in the first camera's coordinate frame.

\medskip
\textbf{Input contract.}

The input must contain at least two valid world-to-camera extrinsics.
The first camera defines the reference frame, with positive horizontal
motion pointing right and positive depth motion pointing forward.

\medskip
\textbf{Option contract.}

The answer choices must contain recognizable left, right, and
forward-motion semantics. Their option letters are parsed from the
current question rather than fixed in the skill.

\medskip
\textbf{Failure guards.}

The skill returns a diagnostic when the extrinsic stack is missing or
malformed, fewer than two cameras are available, the motion magnitude
is degenerate, or the options cannot be parsed.

\end{skillcard}

The core structure of the evolved kernel is shown below. Detailed
validation messages and scoring adjustments are omitted for clarity.

\begin{codecard}{EVOLVED CAMERA-MOTION GEOMETRY KERNEL}
# Geometry skill:
# camera0_motion_option_from_extrinsics

def reusable_decision(extrinsic, options_text):
    ext, error = normalize_extrinsics(
        extrinsic
    )
    if error is not None:
        return diagnostic(error)

    options = parse_motion_options(
        options_text
    )
    if not options:
        return diagnostic(
            "no_recognized_motion_options"
        )

    center_0 = camera_center(ext[0])
    center_1 = camera_center(ext[1])

    motion_world = center_1 - center_0
    motion_cam0 = to_camera0_frame(
        motion_world, ext[0]
    )

    motion_xz = normalize_motion(
        motion_cam0[[0, 2]]
    )
    if motion_xz is None:
        return diagnostic(
            "motion_too_small_or_invalid"
        )

    scores = {
        option["label"]: motion_score(
            motion_xz, option
        )
        for option in options
    }

    return max(scores, key=scores.get)
\end{codecard}

Compared with the original sample-specific programs, the evolved skill
makes the reference frame explicit, parses the current option mapping,
and replaces unsafe default answers with observable diagnostics.

\casestep{Later Retrieval and Reuse.}
The later query presented in
Case 1 again asks for the motion between
two camera views. Its Tool-Use pipeline first invokes
\path{GeometricReconstructor.reconstruct}, producing the required
camera extrinsics.

At the subsequent Python step, the Geometry selector initially sees
the compact description of
\path{camera0_motion_option_from_extrinsics}. After selecting it, the
runtime exposes the complete usage conditions and reusable kernel. The
current reconstruction and answer choices are then bound to the
kernel's inputs.

\begin{skillcard}{Instantiated Evolved Geometry Skill}
\skillid{camera0_motion_option_from_extrinsics}
\textbf{Current bindings.}

Extrinsic input: camera poses returned by the current reconstruction

Reference frame: the first input camera

Compared camera: the second input camera

Current options: direct-left, direct-right, diagonal-forward-left, and
diagonal-forward-right

\medskip
\textbf{Execution.}

The kernel recovers both camera centers, expresses the displacement of
the second camera in the first-camera frame, and compares its lateral
and forward components with the current option semantics.

\medskip
\textbf{Output.}

The recovered motion is diagonally forward and left, producing
option~\textbf{C} for this instance.

\end{skillcard}

The result illustrates the full Geometry-skill evolution cycle:
sample-specific computation is observed in traces, recurring atomic
operations are induced, fusion creates a guarded reusable kernel, and
the resulting skill is retrieved to guide later numerical execution.

\medskip
\noindent
\emph{Evolution path:}
independent generated programs
$\rightarrow$ recurring geometric pattern
$\rightarrow$ atomic-operation induction
$\rightarrow$ guarded Geometry kernel
$\rightarrow$ later retrieval and reuse.

%% file: Appendix/C.tex
\section{Evolved Skill Inventory}
\label{app:evolved-skill-inventory}

The following tables list the skills retained after online evolution.
Initial seed skills are excluded, and readable names replace internal
identifiers.
\vspace{-2pt}

\newcommand{\skillsep}{\cmidrule(l){2-2}}
\begin{table}[H]
\centering
\caption{Evolved skill inventory for MindCube.}
\label{tab:mindcube-evolved-skills}
\footnotesize
\setlength{\tabcolsep}{4pt}
\renewcommand{\arraystretch}{0.98}
\begin{tabularx}{\columnwidth}{
    @{}
    >{\raggedright\arraybackslash}p{0.23\columnwidth}
    >{\raggedright\arraybackslash}X
    @{}
}
\toprule
\rowcolor{cardtitle}
\textbf{Skill type} & \textbf{Skill name} \\
\midrule
\multirow{11}{*}{\bfseries Tool-Use}
& Four-View Behind-Me Camera Layout \\
\skillsep
& Explicit Turn-View Symbolic Layout \\
\skillsep
& Four-View Turn-and-Move Symbolic Layout \\
\skillsep
& Behind-Me Camera Layout with Behind-View Option Grounding \\
\skillsep
& Behind-Me Resolved-View Option Grounding \\
\skillsep
& Query-View Candidate Grounding with a Target-Pose Frame \\
\skillsep
& Four-View Turn-and-Move Target Grounding \\
\skillsep
& Query-View Candidate Grounding with a Query-View Target Pose \\
\skillsep
& Behind-Me Symbolic Layout followed by Option Grounding \\
\skillsep
& Turned-View Target Grounding \\
\skillsep
& Turned-View Target Grounding V2 \\
\midrule
\multirow{3}{*}{\bfseries Geometry}
& Camera-0 Motion from Extrinsics \\
\skillsep
& Right-Turn Distance Change to a Target \\
\skillsep
& Left-Turn Proximity in the Camera Frame \\
\bottomrule
\end{tabularx}
\end{table}

\begin{table}[H]
\centering
\caption{Evolved skill inventory for MMSI.}
\label{tab:mmsi-evolved-skills}
\footnotesize
\setlength{\tabcolsep}{4pt}
\renewcommand{\arraystretch}{0.98}
\begin{tabularx}{\columnwidth}{
    @{}
    >{\raggedright\arraybackslash}p{0.23\columnwidth}
    >{\raggedright\arraybackslash}X
    @{}
}
\toprule
\rowcolor{cardtitle}
\textbf{Skill type} & \textbf{Skill name} \\
\midrule
\multirow{3}{*}{\bfseries Tool-Use}
& Single-Anchor Grounding with Camera-Layout Binding \\
\skillsep
& Reconstruction and Camera-Layout Reasoning for Egomotion \\
\skillsep
& Dual-Landmark Grounding for Observer-Relative Comparison \\
\midrule
\multirow{3}{*}{\bfseries Geometry}
& Object-Side Reasoning from Pose Bearing \\
\skillsep
& Camera-Frame Object-Quadrant Selection \\
\skillsep
& Absolute Compass-Side Reasoning from a Reference View \\
\bottomrule
\end{tabularx}
\end{table}

\begin{table}[H]
\centering
\caption{Evolved skill inventory for OmniSpatial.}
\label{tab:omnispatial-evolved-skills}
\footnotesize
\setlength{\tabcolsep}{4pt}
\renewcommand{\arraystretch}{0.98}
\begin{tabularx}{\columnwidth}{
    @{}
    >{\raggedright\arraybackslash}p{0.23\columnwidth}
    >{\raggedright\arraybackslash}X
    @{}
}
\toprule
\rowcolor{cardtitle}
\textbf{Skill type} & \textbf{Skill name} \\
\midrule
\multirow{10}{*}{\bfseries Tool-Use}
& Refined Entity Redetection after Grounding Review \\
\skillsep
& Camera-Arrival Grounding Review \\
\skillsep
& Detection and Grounding Ambiguity Review \\
\skillsep
& Allocentric Pairwise Grounding-Readiness Review \\
\skillsep
& Egocentric Relation Failure Review \\
\skillsep
& Allocentric Egocentric-Relation Review \\
\skillsep
& Reference-Frame Relation Counting \\
\skillsep
& Grounding-Consistency Review \\
\skillsep
& Viewpoint-Conditioned Region Membership \\
\skillsep
& Allocentric Axial-Relation Review \\
\midrule
\multirow{2}{*}{\bfseries Geometry}
& Camera-Relative Travel-Time Selection \\
\skillsep
& Allocentric Left--Right Reasoning from Reference Pose \\
\bottomrule
\end{tabularx}
\end{table}

%% file: Appendix/D.tex
\section{Prompt Templates}
\label{app:prompt_templates}

This section presents the core prompts used during inference and online skill evolution.
To improve readability, we organize each prompt into four parts:
\textbf{Purpose}, \textbf{Inputs}, \textbf{Instructions}, and \textbf{Output format}.
Full prompts will be available after the code release.

\begin{promptbox}{SKILL RETRIEVAL PROMPT}

\textbf{Purpose.}
Select a small set of reusable rules that are relevant to the current task and workspace context.
The selector sees only \emph{compact} rule descriptions at this stage.

\textbf{Inputs.}
\begin{itemize}[leftmargin=1.5em, itemsep=2pt]
    \item \texttt{<TASK\_INSTRUCTION>}
    \item \texttt{<CONTEXT\_SUMMARY>}
    \item \texttt{<RULE\_TYPE>}
    \item \texttt{<TARGET\_ACTION\_IF\_APPLICABLE>}
    \item \texttt{<SELECTION\_CONTEXT>}
    \item \texttt{<CANDIDATE\_RULES\_JSON>}
    \item \texttt{<MAX\_RULES>}
    \item optional \texttt{<ERROR\_FEEDBACK>}
\end{itemize}

\textbf{Instructions.}
\begin{enumerate}[leftmargin=1.5em, itemsep=3pt]
    \item Select only rules whose \texttt{docstring} fits the current task and context.
    \item Use \texttt{structure\_hint} only to distinguish otherwise similar rules.
    \item Do not select a rule merely because one word or one tool overlaps.
    \item It is valid to select no rules if none are specific enough.
    \item Return at most \texttt{<MAX\_RULES>} rule ids.
\end{enumerate}

\textbf{Strict output format.}
\begin{lstlisting}
{
  "selected_rule_ids": [
    "rule_id_if_useful"
  ],
  "rationale": "brief reason for the selection, or why none apply"
}
\end{lstlisting}

If no rule is useful, return:
\begin{lstlisting}
{
  "selected_rule_ids": [],
  "rationale": "no candidate rule is specific enough for this task"
}
\end{lstlisting}

\end{promptbox}

\begin{promptbox}{INFERENCE PLANNER PROMPT}

\textbf{Purpose.}
Act as the planner for a tool-using spatial reasoning agent.
At each turn, choose exactly one top-level decision:
use one complete Tool-Use skill, execute one immediate tool action, or return the final answer.

\textbf{Inputs.}
\begin{itemize}[leftmargin=1.5em, itemsep=2pt]
    \item \texttt{<TASK\_INSTRUCTION>}
    \item \texttt{<WORKSPACE\_SUMMARY>}
    \item \texttt{<RECENT\_OBSERVATIONS>}
    \item \texttt{<CANDIDATE\_BLOCK\_RULES>}
    \item \texttt{<AVAILABLE\_TOOL\_SCHEMAS>}
    \item optional \texttt{<ERROR\_FEEDBACK>}
\end{itemize}

\textbf{Instructions.}
\begin{enumerate}[leftmargin=1.5em, itemsep=3pt]
    \item Choose exactly one decision type:
    \begin{itemize}[leftmargin=1.5em, itemsep=2pt]
        \item \texttt{decision\_type="rule"}: select one reusable high-level rule by id;
        \item \texttt{decision\_type="action"}: emit one immediate tool action or the final answer.
    \end{itemize}
    \item A block rule is a semantically complete reusable procedure, not an arbitrary tool-call fragment.
    \item The executor does not decide final answers; return a final answer only when workspace evidence is sufficient.
    \item Use \texttt{missing\_information} to list unresolved semantic roles, such as the reference frame, reference entity, target, candidate set, requested relation, or option mapping.
    \item One aggregate detection cannot fill multiple semantic roles.
    \item Prefer one small tool action at a time unless multiple calls are clearly independent.
    \item For spatial direction, distance, or orientation, prefer explicit geometry and deterministic computation.
    \item For an object-centric viewpoint, use object-pose evidence when a visible entity defines the reference orientation.
\end{enumerate}

\textbf{Strict output format.}

Rule decision:
\begin{lstlisting}
{
  "decision_type": "rule",
  "use_rule": true,
  "selected_rule_id": "exact_rule_id",
  "selected_rule_reason": "why this complete rule matches the current need",
  "missing_information": [
    "unresolved semantic role"
  ],
  "tool_calls": [],
  "final_answer": null
}
\end{lstlisting}

Immediate action:
\begin{lstlisting}
{
  "decision_type": "action",
  "use_rule": false,
  "selected_rule_id": null,
  "selected_rule_reason": "why an immediate action is needed",
  "missing_information": [
    "unresolved semantic role"
  ],
  "tool_calls": [
    {
      "tool_name": "Tool.method",
      "output_variable": "workspace_name",
      "args": {}
    }
  ],
  "final_answer": null
}
\end{lstlisting}

Final answer:
\begin{lstlisting}
{
  "decision_type": "action",
  "use_rule": false,
  "selected_rule_id": null,
  "selected_rule_reason": "workspace already satisfies the task",
  "missing_information": [],
  "tool_calls": [],
  "final_answer": "answer derived from workspace evidence"
}
\end{lstlisting}

\end{promptbox}

\begin{promptbox}{TOOL-USE SKILL INDUCTION PROMPT}

\textbf{Purpose.}
Infer one reusable pipeline extension from a batch of compact traces.
The goal is to grow a narrower child pipeline from an existing reusable trunk whenever possible.

\textbf{Inputs.}
\begin{itemize}[leftmargin=1.5em, itemsep=2pt]
    \item \texttt{<ATOM\_CONTRACTS\_JSON>}
    \item \texttt{<CURRENT\_PIPELINE\_KB\_JSON>}
    \item \texttt{<TRACE\_BATCH\_JSON>}
    \item optional \texttt{<ERROR\_FEEDBACK>}
\end{itemize}

\textbf{Instructions.}
\begin{enumerate}[leftmargin=1.5em, itemsep=3pt]
    \item Group traces by task family and required computation.
    \item A valid family must contain at least two traces with the same semantic roles, evidence needs, and decision pattern, including at least one correct trace.
    \item Generate at most one candidate from one supported family.
    \item Use \texttt{operation="branch"} when an existing parent pipeline supplies a reusable trunk.
    \item Record only the missing or replaced semantic operations in \texttt{extension\_steps}; fusion will materialize the complete child.
    \item Use \texttt{operation="new\_root"} only when no reusable parent exists.
    \item Do not generate incomplete fragments such as reconstruction-only, retry-only, or isolated detection/projection steps.
    \item Do not invent tools, and do not place Geometry skill ids inside a Tool-Use pipeline.
    \item The stopping condition must describe a semantically complete and observable workspace state.
\end{enumerate}

\textbf{Strict output format.}
\begin{lstlisting}
{
  "candidate_rules": [
    {
      "id": "short_snake_case_extension",
      "rule_type": "pipeline_delta",
      "operation": "branch",
      "parent_rule_id": "existing_pipeline_id",
      "docstring": "narrow task family and added evidence need",
      "extension_steps": [
        {
          "atom": "Tool.method",
          "instruction": "missing semantic operation",
          "bindings": {
            "required_arg": "$semantic_workspace_slot"
          },
          "output": "semantic_output_name"
        }
      ],
      "stopping_condition": "semantically complete child result"
    }
  ],
  "rationale": "supporting trace family and recurring gap"
}
\end{lstlisting}

If no supported reusable program exists, return:
\begin{lstlisting}
{
  "candidate_rules": [],
  "rationale": "no supported reusable program exists"
}
\end{lstlisting}

\end{promptbox}

\begin{promptbox}{GEOMETRY SKILL INDUCTION PROMPT}

\textbf{Purpose.}
Induce a reusable geometry kernel for \texttt{PythonTool.code} from execution records.
The induced rule should capture a complete recurring computation, together with its applicability conditions and failure guards.

\textbf{Inputs.}
\begin{itemize}[leftmargin=1.5em, itemsep=2pt]
    \item \texttt{<CURRENT\_CODE\_CONTROL\_KB\_JSON>}
    \item \texttt{<PYTHON\_TRACE\_BATCH\_JSON>}
    \item optional \texttt{<ERROR\_FEEDBACK>}
\end{itemize}

\textbf{Instructions.}
\begin{enumerate}[leftmargin=1.5em, itemsep=3pt]
    \item Group records by the same computation, required variables, reference frame, and output decision.
    \item A valid family must contain at least two traces, including at least one correct execution.
    \item Prefer families where correct and incorrect traces expose a concrete reusable difference.
    \item Generalize the complete decision computation rather than a trivial helper or a shared code substring.
    \item \texttt{usage} must specify required variables, shapes, reference-frame assumptions, and per-call bindings.
    \item \texttt{code\_lines} must define one self-contained reusable kernel, possibly with helper functions and one decision function.
    \item Missing or malformed evidence must return a diagnostic rather than a fabricated answer.
    \item Do not encode sample ids, one-off object names, option letters, or sample-specific answers.
\end{enumerate}

\textbf{Strict output format.}
\begin{lstlisting}
{
  "candidate_rules": [
    {
      "id": "short_snake_case_kernel",
      "rule_type": "code_control",
      "level": "high_level",
      "docstring": "narrow coding scenario and evidence conditions",
      "usage": [
        "required inputs, frames, modes, and bindings",
        "conditions under which the kernel must not be used"
      ],
      "code_lines": [
        "def reusable_decision(required_values, mode):",
        "    # complete reusable computation",
        "    return result"
      ]
    }
  ],
  "rationale": "why the traces share one reusable computation"
}
\end{lstlisting}

If unsupported, return:
\begin{lstlisting}
{
  "candidate_rules": [],
  "rationale": "no supported reusable computation exists"
}
\end{lstlisting}

\end{promptbox}

\begin{promptbox}{SKILL FUSION PROMPT}

\textbf{Purpose.}
Fuse the mutable main library and the candidate library into a bounded active library.
For Tool-Use skills, fusion materializes complete child pipelines; for Geometry skills, it merges reusable kernels.

\textbf{Inputs.}
\begin{itemize}[leftmargin=1.5em, itemsep=2pt]
    \item \texttt{<MAX\_RULES>}
    \item \texttt{<READ\_ONLY\_PARENT\_RULES\_JSON>}
    \item \texttt{<MUTABLE\_MAIN\_KB\_JSON>}
    \item \texttt{<CANDIDATE\_KB\_JSON>}
    \item \texttt{<TRACE\_PROOF\_RESERVOIR\_JSON>}
    \item optional \texttt{<ERROR\_FEEDBACK>}
\end{itemize}

\textbf{Instructions.}
\begin{enumerate}[leftmargin=1.5em, itemsep=3pt]
    \item Preserve the schema of Tool-Use pipelines and Geometry kernels.
    \item Parent pipelines are read-only references and must not be rewritten.
    \item For \texttt{pipeline\_delta} candidates with \texttt{operation="branch"}, preserve the reusable parent trunk and insert the extension where it semantically belongs.
    \item Never emit \texttt{pipeline\_delta} directly into the active library; always materialize a complete child.
    \item Merge duplicate children and discard unsupported, overly broad, or sample-specific records.
    \item Discard children that still fail to reach the evidence state claimed by their docstring.
    \item Tool-Use pipelines may reference only native tool atoms and planner reasoning transitions.
    \item When trace-supported fusion is enabled, provide one \texttt{pipeline\_proofs} entry for every new or materially changed pipeline.
\end{enumerate}

\textbf{Strict output format.}
\begin{lstlisting}
{
  "rules": [
    {
      "id": "complete_reusable_skill",
      "rule_type": "pipeline_or_code_control",
      "...": "fields from the corresponding schema"
    }
  ],
  "pipeline_proofs": [
    {
      "rule_id": "materialized_pipeline",
      "segments": [
        {
          "steps": [
            {
              "atom": "Tool.method",
              "input_keys": ["required_arg"],
              "semantic_roles": ["target", "camera"]
            }
          ],
          "prototype_trace_ids": [
            "correct_trace_1",
            "correct_trace_2"
          ],
          "counterexample_trace_ids": [
            "incorrect_trace_1",
            "incorrect_trace_2"
          ]
        }
      ]
    }
  ]
}
\end{lstlisting}

\end{promptbox}

\begin{promptbox}{GEOMETRY-GUIDED CODE GENERATION PROMPT}

\textbf{Purpose.}
Generate the executable wrapper for \texttt{PythonTool.code}.
The code generator receives the current variables, their schemas, the reference-frame requirements, and optionally a retrieved Geometry skill.

\textbf{Inputs.}
\begin{itemize}[leftmargin=1.5em, itemsep=2pt]
    \item \texttt{<USER\_REQUEST>}
    \item \texttt{<REFERENCE\_FRAME\_DESCRIPTION>}
    \item \texttt{<COMPUTATIONAL\_OBJECTIVE>}
    \item \texttt{<CONTEXT\_DESCRIPTION>}
    \item \path{<VARIABLE_SCHEMAS_AND_DOCUMENTATION>}
    \item \texttt{<SELECTED\_CODE\_CONTROL\_DOCSTRING>}
    \item \texttt{<SELECTED\_CODE\_CONTROL\_USAGE>}
    \item \texttt{<SELECTED\_REUSABLE\_KERNEL>}
    \item optional \texttt{<PREVIOUS\_EXECUTION\_ERROR>}
\end{itemize}

\textbf{Instructions.}
\begin{enumerate}[leftmargin=1.5em, itemsep=3pt]
    \item Write one Python function that implements the requested computation.
    \item Use only a reusable kernel whose input contract matches the current variables and objective.
    \item Bind kernel inputs to current workspace values.
    \item Do not redefine the selected helper block.
    \item Adapt only the wrapper, current option mapping, indices, thresholds, and return format.
    \item Verify the computation using the supplied variable documentation.
    \item For multiple-choice questions, evaluate the current options rather than relying on fixed option letters.
    \item Missing or incomparable evidence should result in a diagnostic rather than a fabricated answer.
    \item The returned result must be serializable.
\end{enumerate}

\textbf{Strict output format.}

When a reusable Geometry kernel is selected:
\begin{lstlisting}
{
  "use_code_blocks": [
    "exact_geometry_skill.kernel"
  ],
  "execute_code":
"def execute(current_variables):\n\
    result = reusable_decision(...)\n\
    return result"
}
\end{lstlisting}

Otherwise, return exactly one Python code block defining:
\begin{lstlisting}
def execute(current_variables):
    import ...
    ...
    return serializable_value
\end{lstlisting}

\end{promptbox}